\documentclass[10pt,twocolumn,letterpaper]{article}
\pdfoutput=1
\input{config.sty}
\usepackage{amsmath,amsfonts,bm}
\usepackage{stfloats}

\def\eqref#1{equation~\ref{#1}}

\def\1{\bm{1}}

\DeclareMathOperator*{\argmax}{arg\,max}

\date{}
\begin{document}

\title{On the Complexity of Bayesian Generalization}

\author{
Yu-Zhe Shi$^{1,2\,\star{}\,\textrm{\Letter}}$, Manjie Xu$^{1,2\,\star{}}$, John E. Hopcroft$^6$, Kun He$^5$, Joshua B. Tenenbaum$^4$, \\
Song-Chun Zhu$^{1,2,3}$, Ying Nian Wu$^3$, Wenjuan Han$^{2\,\textrm{\Letter}}$, Yixin Zhu$^{1\,\textrm{\Letter}}$
\vspace{0.5em}\\
\small$^1$Institute for AI, Peking University\quad{}
\small$^2$Beijing Institute for General Artificial Intelligence (BIGAI)\\
\small$^3$Department of Statistics, UCLA\quad{}
\small$^4$Department of Brain and Cognitive Sciences, MIT\\
\small$^5$Department of Computer Science, Huazhong University of Science and Technology\\
\small$^6$Department of Computer Science, Cornell University\\
\vspace{0.5em}\\
\small$^\star{}$Equal contributors\quad
\small$\textrm{\Letter}$\phantom\,\,\texttt{\{shiyuzhe,hanwenjuan\}@bigai.ai, yixin.zhu@pku.edu.cn}
}

\maketitle

\begin{abstract}
We consider concept generalization at a large scale in the diverse and natural visual spectrum. Established computational modes (\ie, rule-based or similarity-based) are primarily studied isolated and focus on confined and abstract problem spaces. In this work, we study these two modes when the problem space scales up, and the \textit{complexity} of concepts becomes diverse.
Specifically, at the \textbf{representational level}, we seek to answer how the complexity varies when a visual concept is mapped to the representation space. Prior psychology literature has shown that two types of complexities (\ie, subjective complexity and visual complexity)~\cite{griffiths2003algorithmic} build an inverted-U relation~\cite{donderi2006visual,sun2021seeing}. Leveraging \ac{roa}, we computationally confirm the following observation: Models use attributes with high \ac{roa} to describe visual concepts, and the description length falls in an inverted-U relation with the increment in visual complexity.
At the \textbf{computational level}, we aim to answer how the complexity of representation affects the shift between the rule- and similarity-based generalization. We hypothesize that category-conditioned visual modeling estimates the co-occurrence frequency between visual and categorical attributes, thus potentially serving as the prior for the natural visual world. Experimental results show that representations with relatively high subjective complexity outperform those with relatively low subjective complexity in the rule-based generalization, while the trend is the opposite in the similarity-based generalization.
\end{abstract}

\section{Introduction}

\emph{What is a cucumber?} One may respond by \emph{a deep green colored slim-long cylinder with trichomes on the surface is a cucumber}, or directly pick a cucumber---\emph{see, something looks like this is a cucumber}. Given either answer as prior knowledge, you can easily identify cucumbers; you may check whether a target meets the rules described in the first answer or judge whether it is similar to the example shown in the second answer. Such capability is concept generalization, and the approaches used to identify cucumber are rule- and similarity-based generalization~\cite{sloman1998similarity,shepard1987toward}, respectively.

\begin{figure}[t!]
    \centering
    \includegraphics[width=\linewidth]{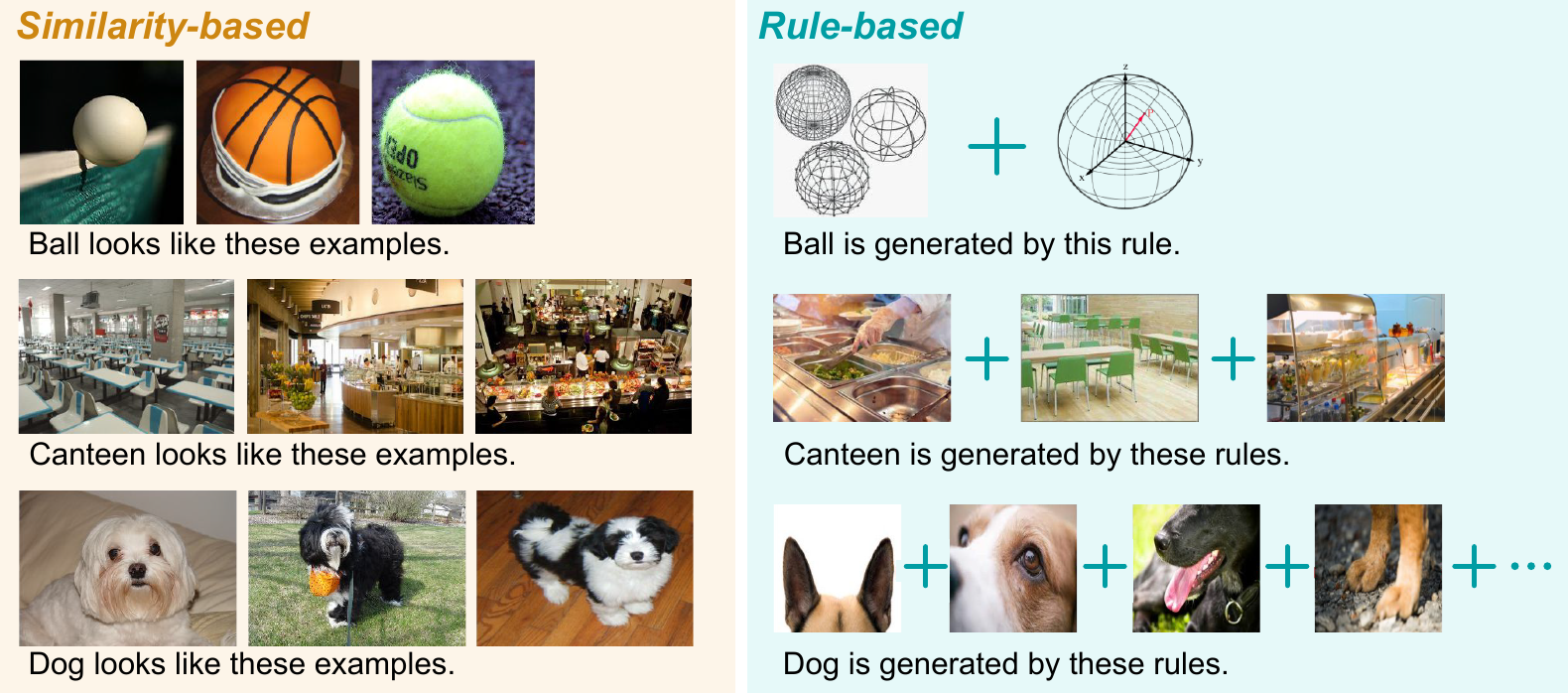}
    \caption{\textbf{Concepts can be described either directly by examples or indirectly by a set of rules.} Here, we show this intuition using the concepts of \emph{ball}, \emph{canteen}, and \emph{dog}; see details in Intro.}
    \label{fig:intro}
\end{figure}

Can both approaches be always applied to all concepts we see in the world? Hardly. Let us consider how people learn to identify \emph{dog}, \emph{canteen}, and \emph{ball}. People easily capture the main feature of a dog given very few examples, yet may get confused with the complex rules to identify a dog; by contrast, people may easily tell the limited rules that shape the concept of canteens, such as serving windows, tables, and chairs; and the concept of the ball is such a simple concept that can be easily captured by examples or a single rule. We refer the readers to \cref{fig:intro} for an illustration. This observation naturally leads to a hypothesis that \emph{whether people generalize through rules or similarity} has something to do with \emph{how complex the concept instances look like}. ``Look like," complexity, and generalization---these three elements shape the hypothesis. In this work, we look into these dimensions of concept generalization.

\begin{figure*}[t!]
    \centering
    \includegraphics[width=0.7\linewidth]{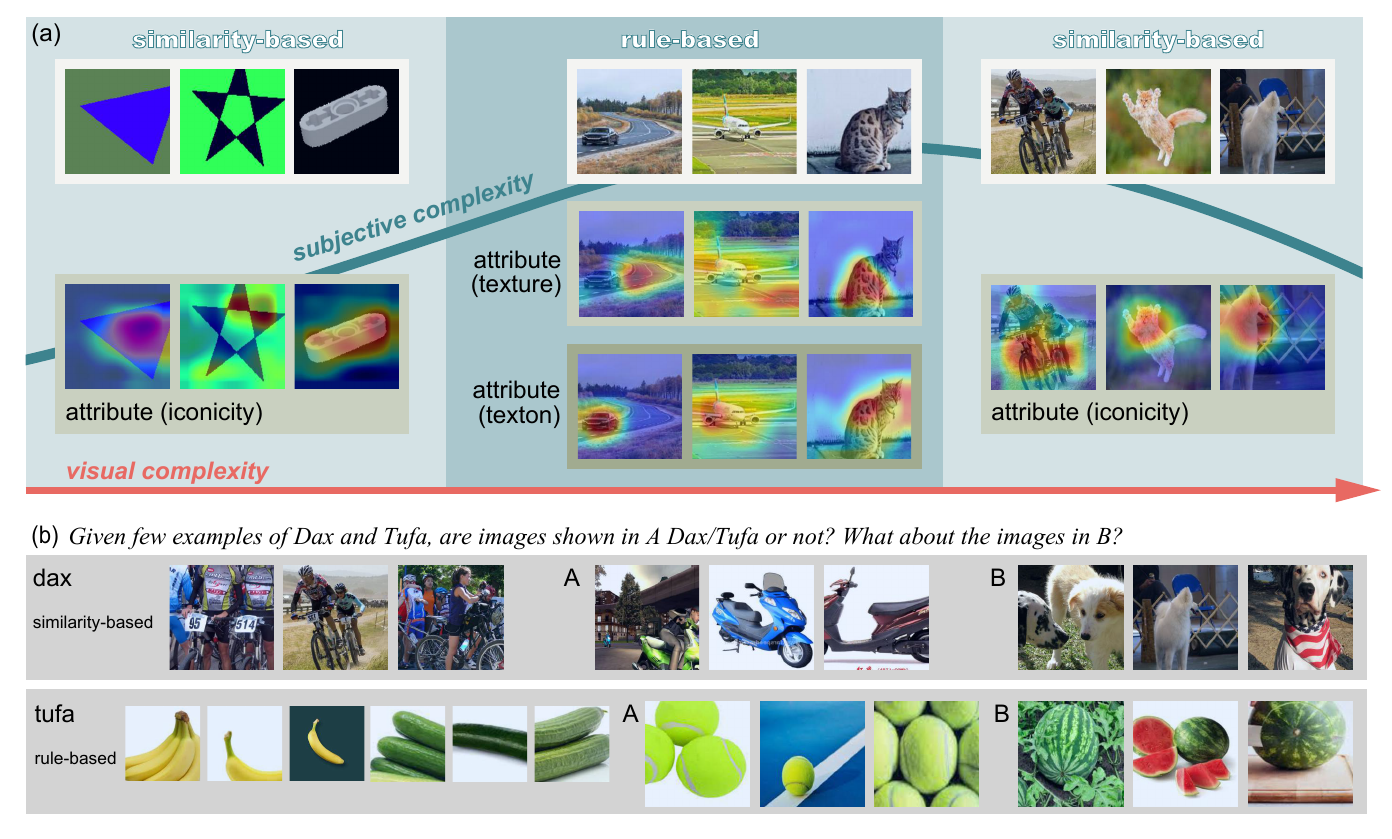}
    \caption{\textbf{The landscape of the computation-mode-shift \vs the concept complexity.} (a) \textit{Representation level:} original visual concepts of diverse complexity and visualization of their representative attributes (around the peaks of heatmaps). (b) \textit{Computation level:} an illustration of similarity- and rule-based generalization. The former is similar to word learning~\cite{xu2007word}: Given very few examples of known concept \textit{dax}, tell which is most likely to be \textit{dax} in unseen examples. The latter is akin to concept learning~\cite{salakhutdinov2012one}: Given a rule \textit{tufa} over two known concepts, tell how \textit{tufa} generates the examples of unknown concepts. As \textcolor{viscplx}{concept visual complexity} increases, \textcolor{subcplx}{concept subjective complexity} first increases, then decreases---and the computation mode shifts from similarity to rules as \textcolor{subcplx}{subjective complexity} increases.}
    \label{fig:overview}
\end{figure*}

First, we contextualize the problem in the literature of concept generalization---the framework of Bayesian inference unifies rule- and similarity-based generalization~\cite{tenenbaum1999rules,tenenbaum1998bayesian,tenenbaum2001generalization,xu2007word}. Based on perception~\cite{kersten2004object}, this paradigm reconstructs human's hypothesis space consisting of abstract features and incubates modern concept learning algorithms~\cite{tenenbaum2011grow,lake2015human,ellis2020dreamcoder}. However, as most concept learners have only demonstrated in confined and abstract problem space, a challenging problem remains: When the problem space \textit{scales up} (\eg, using data collected from the natural world), is there a unified concept representation that combines the two established modes (\ie, rule- and similarity-based)? If it does, how does the generalization shift between the two modes \wrt the \textit{complexity} of concepts?

One concrete hypothesis rooted in psychology~\cite{donderi2006visual,wolfram2002new} is that we tend to describe visual concepts (\ie, \textbf{visual complexity}, the complexity coded by pixels) by simple visual patterns with explicit semantics (\ie, \textbf{subjective complexity}, the coding length for describing certain concepts). For a simple concept, we may only need one attribute (\eg, the shape \emph{circle} for \emph{ball}). As the concepts become more complex, we adopt more attributes, such as \textit{canteens are rooms with serving windows, tables, and chairs}. When the concept becomes even more complicated, we would choose not to describe it---if we still describe it with the attributes generated by the complex rules to identify it, the description would be much too long to be appropriate for communication. Hence, we capture the main feature and view it as an ``icon'' for the concept, such as \textit{dog looks like dog}.

Together, we observe a shift in the continuous space spanned by the rule- and similarity-based approaches \wrt the increase of concept complexity. Intuitively, both very simple and complex concepts have a lower description length, generalized by similarity. In comparison, concepts neither too simple nor too complex have a higher description length, generalized by rules. This observation echos modern literature in both information theory and psychology, which demonstrate that subjective and visual complexity~\cite{griffiths2003algorithmic} come in an inverted-U relation~\cite{donderi2006visual,sun2021seeing}.

In essence, we seek to quantify the relation between the prior-studied but mostly isolated modes (\ie, rule- and similarity-based): What are the \textit{relations} between the computation-mode-shift and the concept complexity, as we hypothesized above and illustrated in \cref{fig:overview}? Specifically, we disassemble the above question into two on the basis of Marr's~\cite{marr1982vision} \textit{representational level} and \textit{computational level}, respectively: (i) How does the complexity change when a visual concept is mapped to the representation space? (ii) How does the complexity of representation affect the shift between rule- and similarity-based generalization? By answering these two questions, we hope to provide a new perspective and the very first pieces of evidence on unifying the two computational modes by mapping out the landscape of the concept complexity \vs the computation mode.

\paragraph{Representation \vs complexity}

Representing the natural visual world merely with human prior is insufficient~\cite{griffiths2016exploring} and oftentimes brittle to generalize. Despite that hierarchy empowers large-scale Bayesian word learning~\cite{miller1998wordnet,abbott2012constructing}, extending it to visual domains is yet challenging and may need costly elaboration. In comparison, modern discriminative models trained for visual categorization by leveraging large-scale datasets can capture the rich concept of attributes~\cite{xie2020representation}. These observations and progresses naturally lead to the problem of concept representation complexity: If we distinguish visual concepts using attributes, at least how many attributes should we use~\cite{chaitin1977algorithmic,li2008introduction}?

To tackle this problem, here we offer a new perspective by bridging the subjective complexity with the visual complexity via \acf{roa}, which consists of (i) the probability of recalling an attribute $z$ when referring to a concept $c$, and (ii) the probability of recalling other concepts $\hat{c}$ when referring to an attribute $z$. This design echoes the principles in rational analysis~\cite{tenenbaum2001rational} yet can be obtained by frequentist statistics for large problem spaces (\eg, natural visual world~\cite{abbott2011testing}).

\paragraph{Computation \vs complexity}

Modern statistical learning methods have demonstrated strong expressiveness in concept representation by implicitly calculating the co-occurrence frequency between visual attributes and categories~\cite{wu2019tale,xie2016theory}, even when scaling up to the complex and large-scale visual domain---the learned representation fits the prior distribution of visual concepts conditioned on categorical description~\cite{xie2020representation}. It can also bridge sensory-derived and language-derived knowledge~\cite{bi2021dual}. Hence, this learning paradigm should somehow have inherent semantic properties in addition to visual properties, such as iconicity~\cite{fay2010interactive,fay2013bootstrap,qiu2022emergent} and disentanglement~\cite{allen2019analogies,gittens2017skip,mikolov2013distributed}.

To properly evaluate the computation, we extend the problem domain from generalization over single concepts to that across multiple concepts. This is because in the natural visual world, we cannot precisely answer how a concept is generated by rules, or which examples are sufficient to represent a concept. Hence, instead of considering the absolute measurements for single concepts, we consider the relative measures between concepts; for example, \emph{cucumber to banana is watermelon to what}, or \emph{dog is more similar to cat or to bike}--only the significant differences are considered. We argue that \textit{rule-based} and \textit{similarity-based} generalization reflects the \textit{analogy} and \textit{similarity} properties in psycholinguistics~\cite{gentner1997structure}, where the former pairs are two ends of a continuum of concept representation, and the latter pairs are two ends of a continuum of literal meaning. Visual categorization brings these two pairs together because linguistic analogy and similarity come from generalizing the corresponding appearance instead of pure literal meaning---concepts with more easy-to-disentangle attributes (\eg, shape and color) are more likely to generalize by rules, while concepts represented with more iconicity~\cite{fay2014iconicity} (\ie, those more likely to be viewed holistically) tend to generalize by similarity.

Computationally, the above hypothesis is consistent with the findings by Wu \etal~\cite{wu2008information}. Specifically in their visual space of natural scenes, textons (low-entropy)~\cite{zhu2005textons} can be composed by very simple concepts~\cite{wu2010learning}, akin to rule-based generalization. In comparison, textures (high-entropy)~\cite{julesz1962visual} cannot be represented by rules~\cite{zhu1997minimax}; instead, they are evaluated and generalized in terms of similarity by ``pursuit''~\cite{zhu1998filters}. As such, we hypothesize that generalization shifts from similarity to rules as subjective complexity increases. Those perspectives provide us with approaches to model and evaluate Bayesian generalization in the natural visual world.

In the remainder of the paper, we first present the new metrics, \acf{roa}, to measure the subjective complexity and analyze the computation-mode-shift in \cref{sec:formulation}. Next, through a series of experiments, we provide strong evidence to support our hypotheses in \cref{sec:exp}; we draw the following \textbf{conclusions} in response to the two problems raised at the beginning:
(i) \textbf{Representation:} the subjective complexity significantly falls in an inverted-U relation with the increment of visual complexity.
(ii) \textbf{Computation:} rule-based generalization is significantly positively correlated with the subjective complexity of the representation, while the trend is the opposite for similarity-based generalization.

\section{Bayesian generalization and complexities}\label{sec:formulation}

In this section, we formulate Bayesian generalization for visual concept learning (\cref{sec:beyasian_generalization}), followed by the definitions of subjective complexity and visual complexity (\cref{sec:complexity}).

\subsection{Bayesian generalization for large-scale visual concept learning}\label{sec:beyasian_generalization}

\paragraph{Concept-conditional modeling}

Let us consider $f:\mathbb{R}^D\mapsto\mathbb{R}^d$, which maps the input $\mathbf{x}\in\mathbb{R}^D$ to a representation vector $\mathbf{z}\in\mathbb{R}^d$. Here, $f$ might be part of a discriminative model trained for visual categorization tasks, such as a prefix for a convolutional neural network without the last fully-connected layer for mapping $\mathbf{z}$ to the category vector $\mathbf{c}\in\mathbb{R}^c$. Training a discriminator for image categorization is to estimate the likelihood of concept $c$ given a set of samples $X$: $P(c|X)=\prod_{x\in X}P(c|x;\theta)$, where $\theta$ is the parameter of $f$. Here, we assume that $f$ provides a good estimation of $P(z|X;\theta)$; Tishby \etal~\cite{tishby2015deep} provides empirical evidence that a discriminative model may first learn how to extract proper attributes to model images $X$ conditioned on $c$, then learn to discriminate their categories based on the attribute distribution. Some dimensions of $\mathbf{z}$ (usually $5\%\sim10\%$ of the total dimensions) capture concrete semantic attributes of visual concepts when the activation score $f_z(X)$ is relatively high~\cite{bau2020understanding}. Combining this concept-conditional measurement with attribute modeling, we rewrite the category prediction considering the attribute as a latent variable and marginalize the observable joint distribution $(X,c)$ over $z$:
\begin{equation}
    \small
    P(c|X;\theta)=\sum_{z\in\mathcal{Z}} P(c,z|X;\theta)=\sum_{z\in\mathcal{Z}} P(c|z)P(z|X;\theta),
\end{equation}
where $\mathcal{Z}$ is the space of all attributes. This expression is essentially a Bayesian prediction view of visual categorization, which can be derived to Bayesian generalization in the natural visual world.

\paragraph{\acf{roa} as an informative prior}

Statistically, we treat the concept-conditional attribute activation score as an estimation of the probability $P(z|c)$ that recalls an attribute $z$ when referring to a concept $c$, similar to answering \textit{``Describe how a dog looks like.''} In the context of the natural visual world, we also have all activation scores generated by an attribute as an estimation of the probability $P(\hat{c}|z)$ that recalls all concepts $\hat{c}\neq c$ when referring to the attribute $z$, akin to answering \textit{``What do you recall seeing a blue thing in a ball shape?''}

Given the above observations and inspiration by Tenenbaum \etal~\cite{tenenbaum2001rational}), we formally define the \ac{roa} of a specific attribute $z_i$ for concept $c$ as:
\begin{equation}
    \ac{roa}(z_i,c)=\log\frac{P(z_i|c)}{\sum_{\hat{c}\neq c}P(\hat{c})P(z_i|\hat{c})},
\end{equation}
where $P(\hat{c})$ is the prior of concepts in the context. We hypothesize that humans estimate $P(\hat{c})$ through both language derivation and visual experience, essentially calculating the co-occurrence frequency between visual attributes and categorical attributes over the joint distribution $P(z_i,\hat{c})$. Hence, modeling \ac{roa} with large-scale image datasets and language corpus should yield human-level prior modeling. On this basis, we use $f$ to statistically estimate $P(z_i,\hat{c})$~\cite{xie2020representation}:
\begin{equation}
    \resizebox{0.9\linewidth}{!}
    {%
        $\displaystyle
        \ac{roa}(z_i,c)=\log\frac{P(z_i|c)}{\sum_{\hat{c}\neq c}P(\hat{c})P(z_i|\hat{c})}\propto\log\frac{P(z_i|c;\theta)}{\sum_{\hat{c}\neq c} P(\hat{c}|z_i;\theta)},
        $
    }
\end{equation}
where $P(z|c;\theta)$ and $P(\hat{c}|z;\theta)$ are estimations of $P(z_i|c)$ and $P(z_i,\hat{c})$, respectively.

\paragraph{Generalize to the unseen}

Given an appropriate modeling of $P(z|X;\theta)$, the goal is to generalize an unknown concept $c^{\prime}$ to a small set of unseen examples $\hat{X}=\{x_1,\cdots,x_n\}$, where $n$ tends to be a small integer. The generalization function $P(c^{\prime}|\hat{X})$ is given by:   
\begin{equation}
    \small
    \begin{aligned}
        P(c^{\prime}|\hat{X})&=\sum_{z\in\mathcal{Z}}P(c^{\prime}|z)P(z|\hat{X};\theta)\\&=\sum_{z\in\mathcal{Z}}\frac{P(c^{\prime})P(z|c^{\prime})}{\sum_{c\in\mathcal{C}}P(c)P(z|c)}P(z|\hat{X};\theta)\\
        &\propto \underset{\substack{\text{uninformative}\\\text{ prior}}}{\underbrace{P(c^{\prime})}} \sum_{z\in\mathcal{Z}} \underset{\substack{\text{informative}\\\text{ prior}}}{\underbrace{\exp\left(\ac{roa}(z,c^{\prime})\right)}} P(z|\hat{X};\theta),
    \end{aligned}
\end{equation}
where the uninformative prior $P(c^{\prime})$ encodes the computation-mode-shift. Specifically, the similarity-based generalization $\langle c::c^{\prime}\rangle$ between a concept pair is defined as $\exists c\in\mathcal{C},\sigma_0(c,c^{\prime})<\delta$, where $\delta$ is a relative small neighbour. Similarly, the rule-based generalization $\langle c_1:c_2::c_3:c^{\prime}\rangle$ over a quadruple of concepts is defined as $\exists c_1,c_2,c_3\in\mathcal{C},\sigma_1(c_1-c_2,c_3-c^{\prime})<\delta$, where $\sigma_{N}(\cdot,\cdot)$ is an arbitrary metric measurement with an N-order input. Further, we define $P(c^{\prime})\propto \sigma_0\sigma_1/(\sigma_0+\sigma_1)$~\cite{tenenbaum1999rules}, resulting in the simplest hypotheses of concepts: The harmonic property keeps guide to similarity-based generalization if $\sigma_0$ is dominating, and vice versa.

\begin{figure}[ht!]
    \centering
    \includegraphics[width=0.65\linewidth]{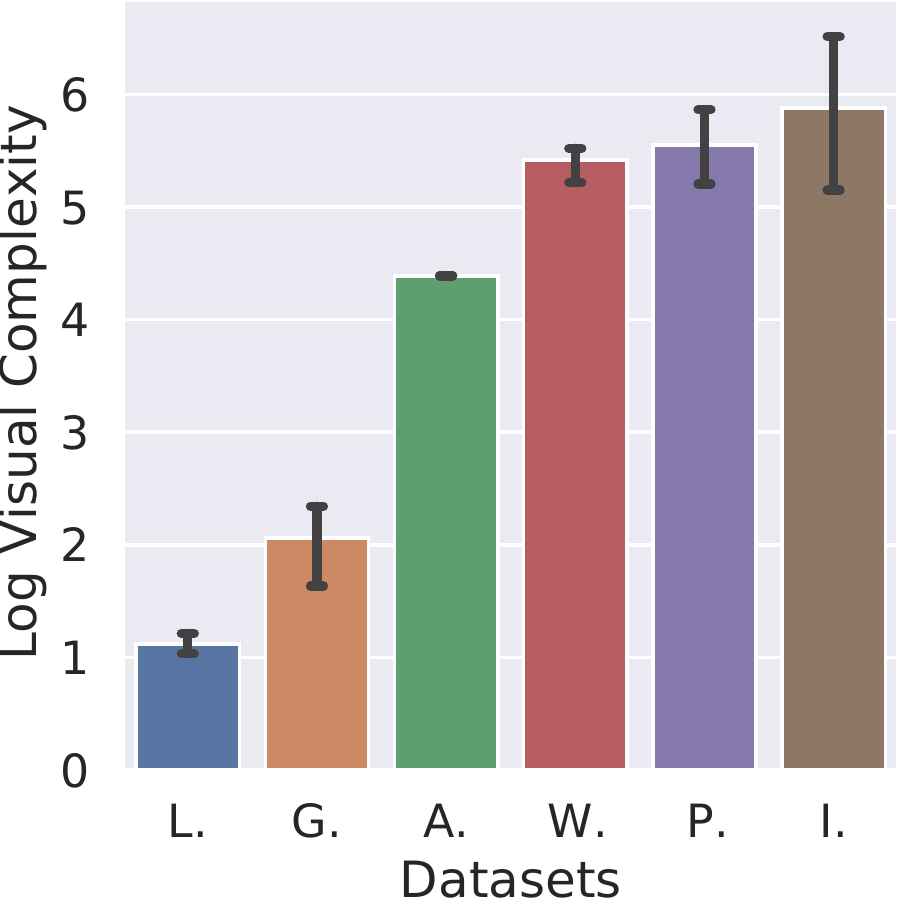}
    \caption{\textbf{Visual complexity of datasets, sorted in increasing order.} L: LEGO, G:2D-Geo, A: ACRE, P: Place, I: ImageNet.}
    \label{fig:dataset_vcplx}
\end{figure}

\subsection{Complexities}\label{sec:complexity}

\paragraph{Visual complexity}

Visual concepts come with diverse complexity, from very simple geometry concepts such as \textit{squares} and \textit{triangles} to very complex natural concepts such as \textit{dogs} and \textit{cats}. Inspired by Wu \etal~\cite{wu2008information}, we indicate concept-wise visual complexity by Shannon's information entropy~\cite{shannon1948mathematical}. Formally, for a set of images $X=\{x_1,x_2,\cdots\}$ belonging to a concept $c$, the concept-wise entropy is $ H(X|c)=\mathbb{E}_{X\sim P(\cdot|c)} [\log P(X|c)]$. As shown in \cref{fig:dataset_vcplx}, we compute the visual complexity and order some commonly known image datasets: 2D geometries~\cite{el20202d}, single concepts~\cite{gill2022fruit}, compositional-attribute objects~\cite{zhang2021acre,johnson2017clevr}, human-made objects~\cite{deng2009imagenet}, scenes~\cite{zhou2017places}, and animals~\cite{xian2018zero,deng2009imagenet}.

\paragraph{Subjective complexity}

We quantify the subjective complexity over the prior model by Kolmogorov Complexity~\cite{li2008introduction}. We calculate the minimum description length, \ie, the minimum number of attributes to discriminate a concept. Specifically, for each concept $c$, we rank all attributes $z\in\mathcal{Z}$ by $\ac{roa}(z,c)$ decreasingly, such that $\forall i,j\in [1,d], i<j, \ac{roa}(z_i,c)\geq \ac{roa}(z_j,c)$. Starting from $K=1$, for each iteration, we select the top-$K$ attributes and check whether these attributes can distinguish the concept $c$ from the others. This process continues if the current iteration cannot distinguish it from the others. Formally, we define subjective complexity of visual concept $L(\hat{c})$ as:
\begin{equation}\label{eq:cplxmeasure}
    \resizebox{0.9\linewidth}{!}{%
        $\displaystyle
        L(\hat{c})=\min\limits_{K}\;\mathbb{1} \big(P(\hat{c}\neq c)<\epsilon \mid c=\argmax\limits_{c}\; P(c|z_1,\cdots,z_K;\phi)\big),
        $}%
\end{equation}
where $\epsilon$ is the error rate threshold, and $\phi$ the parameter of $f$'s suffix in the same discriminative model for visual categorization (\eg, the fully-connected layer). We calculate $P(c|z_1,\cdots,z_K;\phi)$ by removing the neurons' effects corresponding to $z_{K+1},\cdots,z_d$~\cite{bau2020understanding}. Instead of maintaining all error rate thresholds, we leverage the \textit{accuracy gain} between every two iterations to search for the minimum $K$. This process yields the concept-wise subjective complexity in \ac{roa}.

\section{Empirical analysis}\label{sec:exp}

In this section, we provide evidence and analyses to validate the above hypotheses. We (i) conduct empirical analyses at both the \textit{representation} (\cref{sec:representation}) and \textit{computational} (\cref{sec:computation}) level; (ii) provide quantitative analyses of the computation-mode-shift \wrt the concept complexity in \cref{sec:computation}); and (iii) provide qualitative analyses to interpret from the aspect of natural image statistics in \cref{sec:statinterp}.

\begin{figure*}[t!]
    \centering
    \begin{subfigure}[t]{0.9\linewidth}
        \centering
        \includegraphics[width=0.97\linewidth]{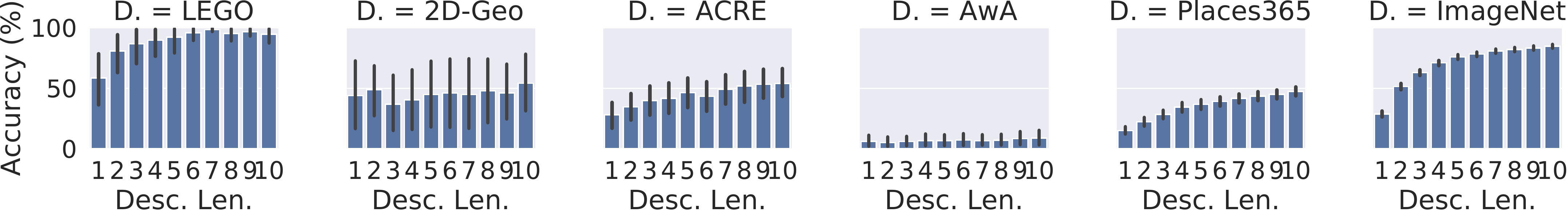}
        \caption{The accuracy of visual categorization task with description length from 1 to 12}
        \label{fig:kacctop}
    \end{subfigure}%
    \\
    \begin{subfigure}[t]{0.33\linewidth}
        \centering
        \includegraphics[width=0.8\linewidth, height=0.58\linewidth]{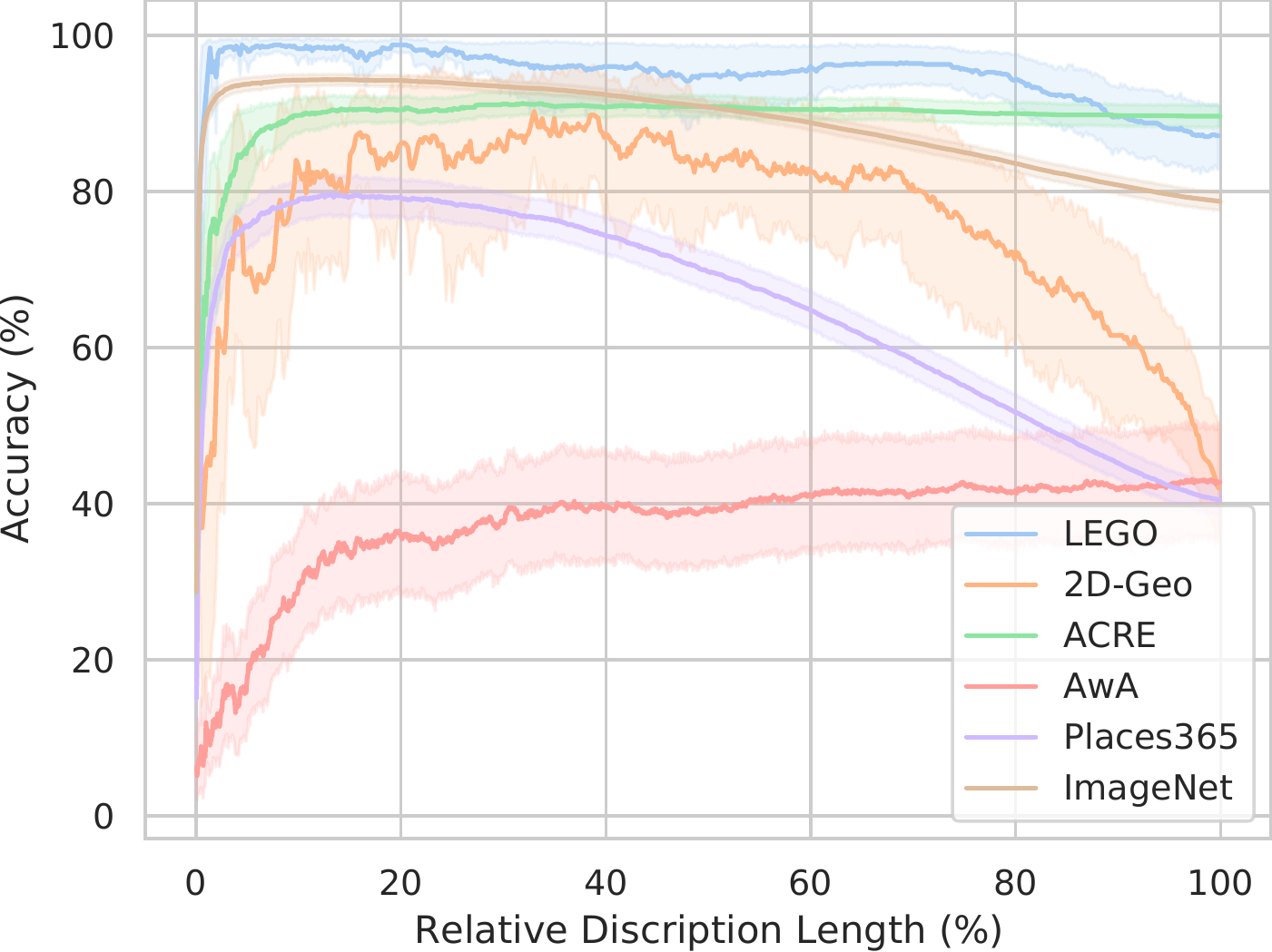}
        \caption{Visual categorization accuracy with various description lengths}
        \label{fig:kcplxaccrate}
    \end{subfigure}%
    \hfill%
    \begin{subfigure}[t]{0.33\linewidth}
        \centering
        \includegraphics[width=0.8\linewidth, height=0.58\linewidth]{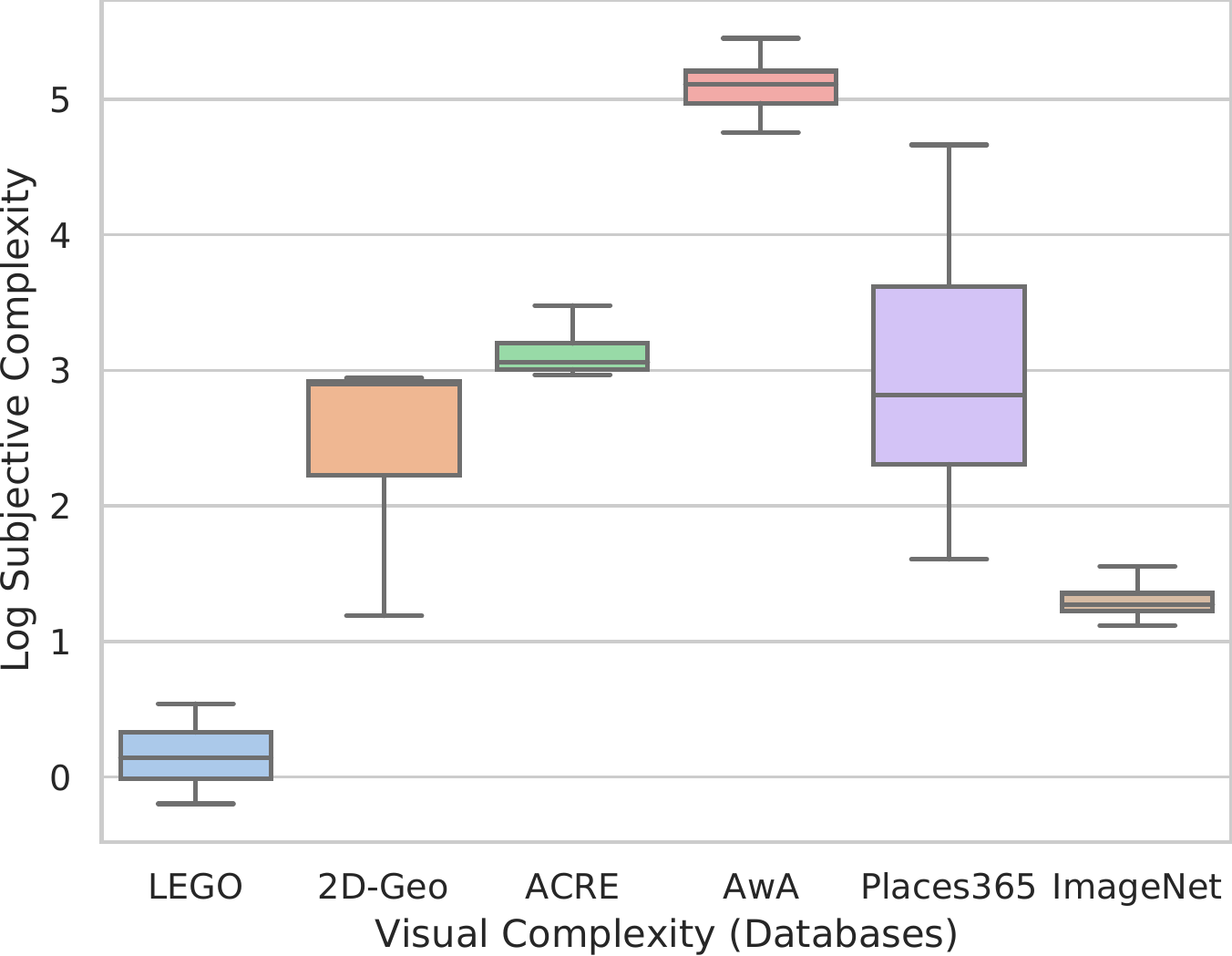}
        \caption{Average subjective complexity on different datasets}
        \label{fig:kcplx_box}
    \end{subfigure}%
    \hfill%
    \begin{subfigure}[t]{0.33\linewidth}
        \centering
        \includegraphics[width=0.8\linewidth, height=0.58\linewidth]{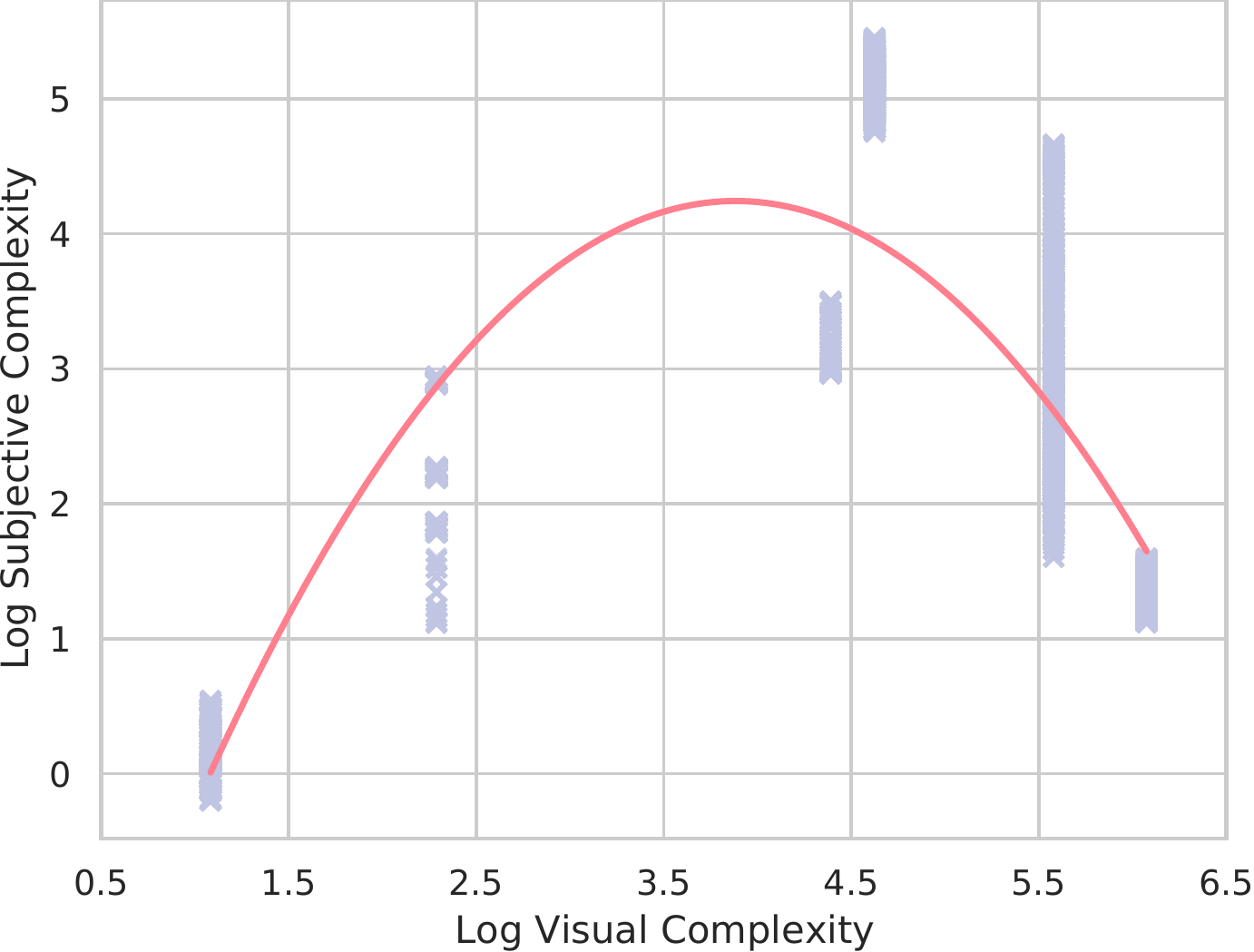}
        \caption{\fontsize{8}{10}\selectfont{}Estimated inverted-U relation between visual and subjective complexity}
        \label{fig:kcplx_lm}
    \end{subfigure}%
    \caption{\textbf{Quantitative results of \textit{Representation \vs Complexity}.} (vector graphics; zoom for details)}
    \label{fig:repsentation}
\end{figure*}

\subsection{Representation \texorpdfstring{\vs}{} complexity}\label{sec:representation}

This experiment investigates the visual concepts' subjective complexity by visual categorization. Our predictions were that models use attributes with high \ac{roa} to describe visual concepts, and the description length falls in an inverted-U relation with the increment of visual complexity.

\paragraph{Method}

Six groups of discriminative models are trained from scratch on six datasets with the supervision of concept labels: LEGO~\cite{tatman2017lego}, 2D-Geo~\cite{el20202d}, ACRE~\cite{zhang2021acre}, AwA~\cite{xian2018zero}, Places~\cite{zhou2017places}, and ImageNet~\cite{deng2009imagenet}, ordered as the increment of concept-wise visual complexity. Models are all optimized to converge on the training set and are tuned to the best hyper-parameters on the validation set. Readers can refer to the supplementary material for details about these datasets and the training process.

During the evaluation, \ac{roa} is calculated for each attribute in the context of all concepts for each dataset. Following the protocol described in \cref{sec:complexity}, the models conduct visual categorization tasks from leveraging only one attribute with the highest \ac{roa} to the entire attribute space.

\paragraph{Results} 

The main quantitative results are illustrated in \cref{fig:repsentation}. Subjective complexity shows significant diversity between the datasets. The logarithm values are as follows; see \cref{fig:kcplx_box}. 
LEGO: $.10\, (CI=[-.10, .52]$, $p<.05)$, 2D-Geo: $2.91\, (CI=[1.21, 2.95]$, $p<.05)$, ACRE: $3.08\, (CI=[2.99,3.46]$, $p<.05)$, AwA: $5.08\, (CI=[4.82, 5.36]$, $p<.05)$, Places: $2.74\, (CI=[1.63,4.72]$, $p<.05)$, ImageNet: $1.28\, (CI=[1.16,1.51]$, $p<.05)$.
All models rely on only a few (less than $20\%$ of all) attributes to reach the prediction accuracy comparable with prediction accuracy exploiting all attributes; see \cref{fig:kcplxaccrate}. Most models (5 out of 6) exploit very few (less than $5\%$ of all) attributes to reach a higher accuracy than that of all attributes; see \cref{fig:kcplxaccrate}. The models for the simplest dataset (\ie, LEGO) and the most complex dataset (\ie, ImageNet) obtain a large accuracy gain (over $10\%$) with the description length from 0 to 3 and obtain smaller accuracy subsequently. In comparison, the models for ACRE and Places obtain relatively small accuracy gain (about $5\%$) with description length from 0 to 8; see \cref{fig:kacctop}. \cref{fig:kcplx_lm} shows the estimated inverted-U relation between subjective complexity and visual complexity. Following the ``two-lines'' test~\cite{simonsohn2018two}, the relation is relatively robust across the datasets, decomposing the non-monotonic relation via a ``breakpoint''; the positive linear relation ($b=1.10,z=253.76, p<1e-4$) and the negative linear relation ($b=-2.57,z=-659.26,p<1e-4$) are both significant.

\paragraph{Discussion} 

The above results reveal that (i) the representation helps the models to describe concepts with very few attributes, (ii) representation trained from very simple or very complex datasets usually have a shorter concept description length than those trained on other datasets, and (iii) the subjective complexity significantly comes in an inverted-U relation with the visual complexity.

\subsection{Computation \texorpdfstring{\vs}{} complexity}\label{sec:computation}

\begin{figure}[b!]
    \centering
    \captionsetup[subfigure]{labelformat=empty}
    \begin{subfigure}[t]{0.409\linewidth}
        \centering
        \includegraphics[width=\linewidth]{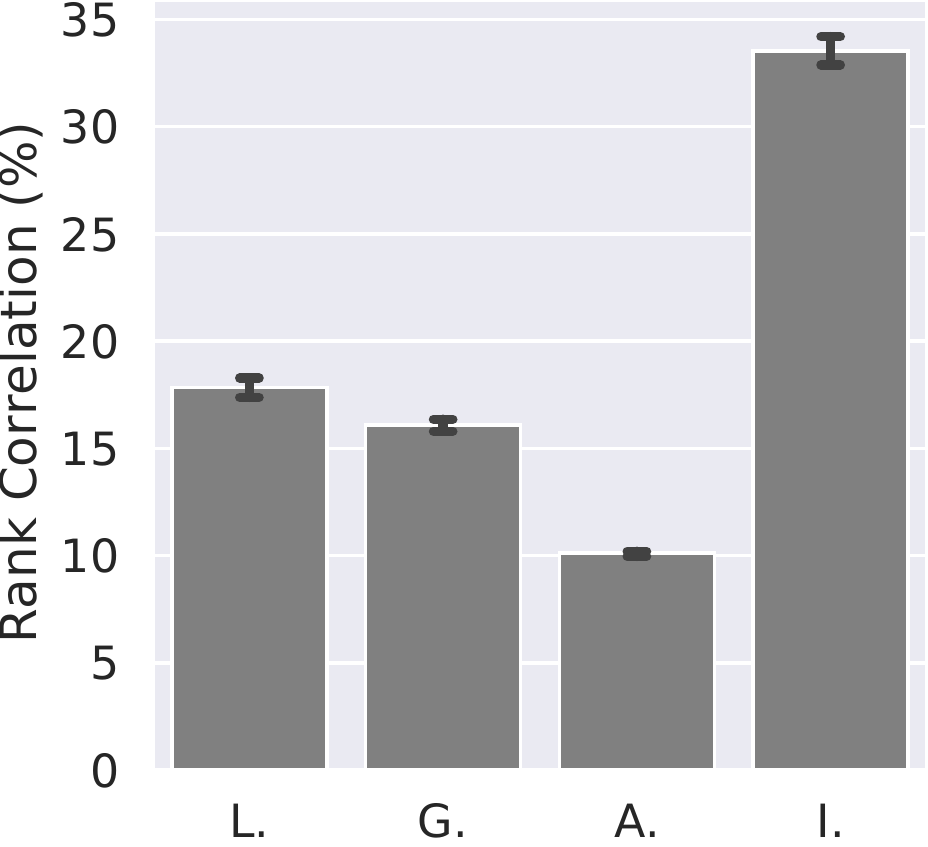}
        \caption{(a)}
        \label{fig:genacc_s}
    \end{subfigure}%
    \begin{subfigure}[t]{0.391\linewidth}
        \centering
        \includegraphics[width=\linewidth]{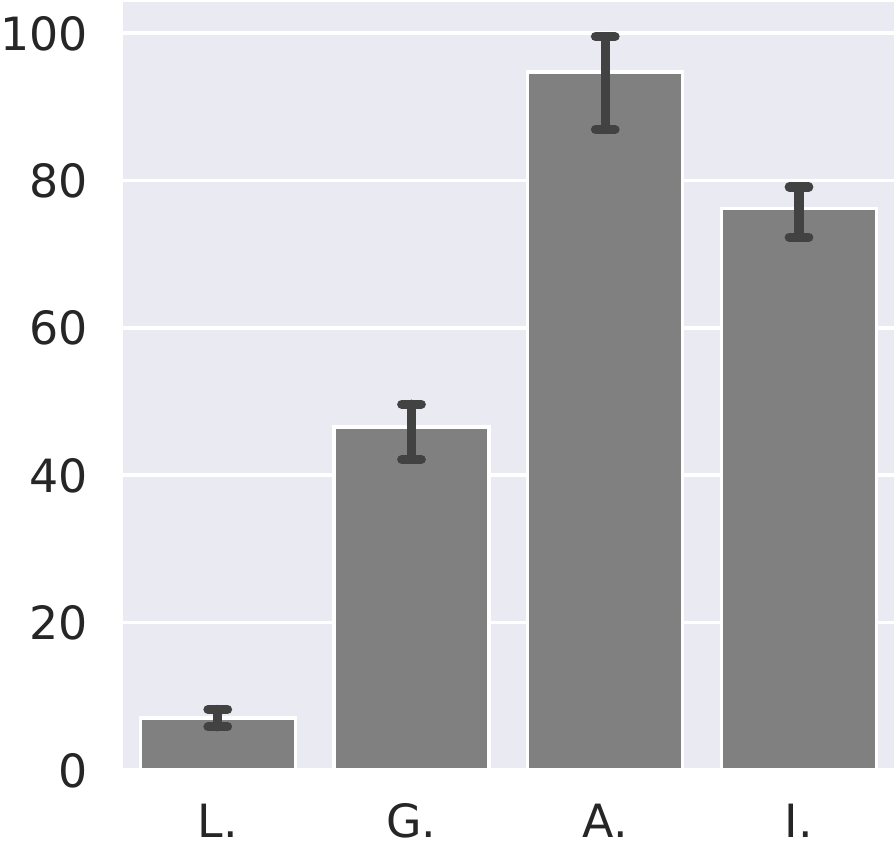}
        \caption{(b)}
        \label{fig:genacc_r}
    \end{subfigure}\\
    \begin{subfigure}[t]{0.409\linewidth}
        \centering
        \includegraphics[width=\linewidth]{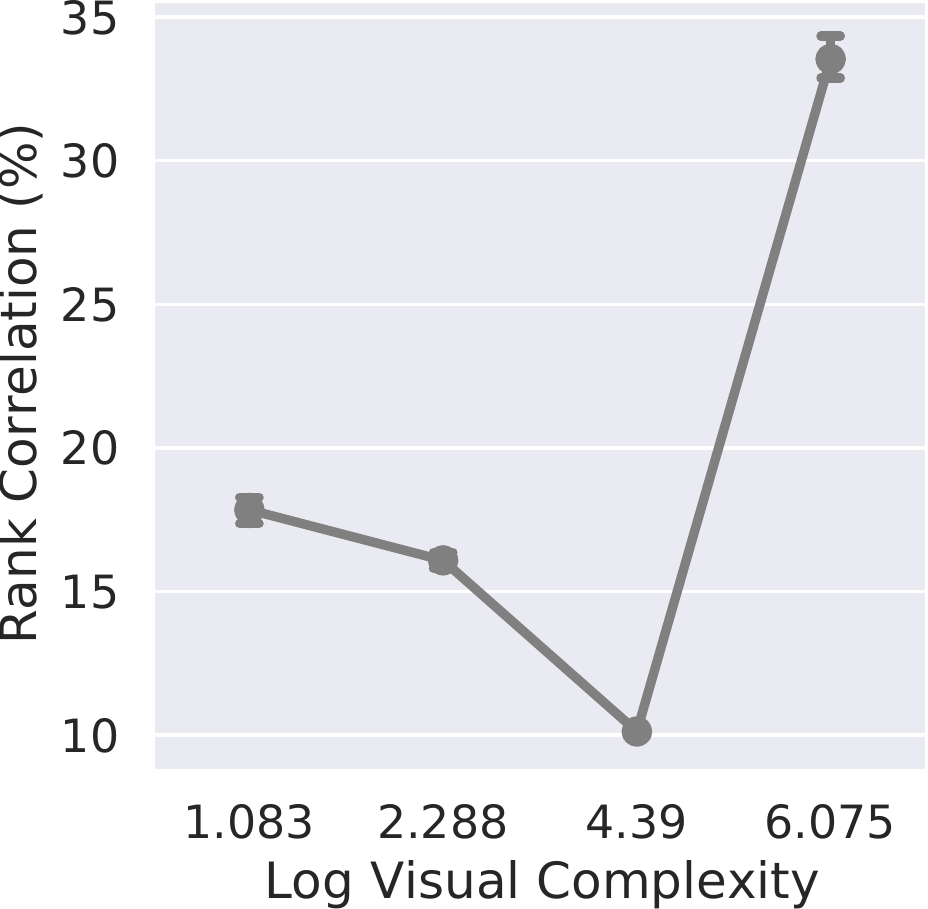}
        \caption{(c)}
        \label{fig:genaccp_s}
    \end{subfigure}%
    \begin{subfigure}[t]{0.391\linewidth}
        \centering
        \includegraphics[width=\linewidth]{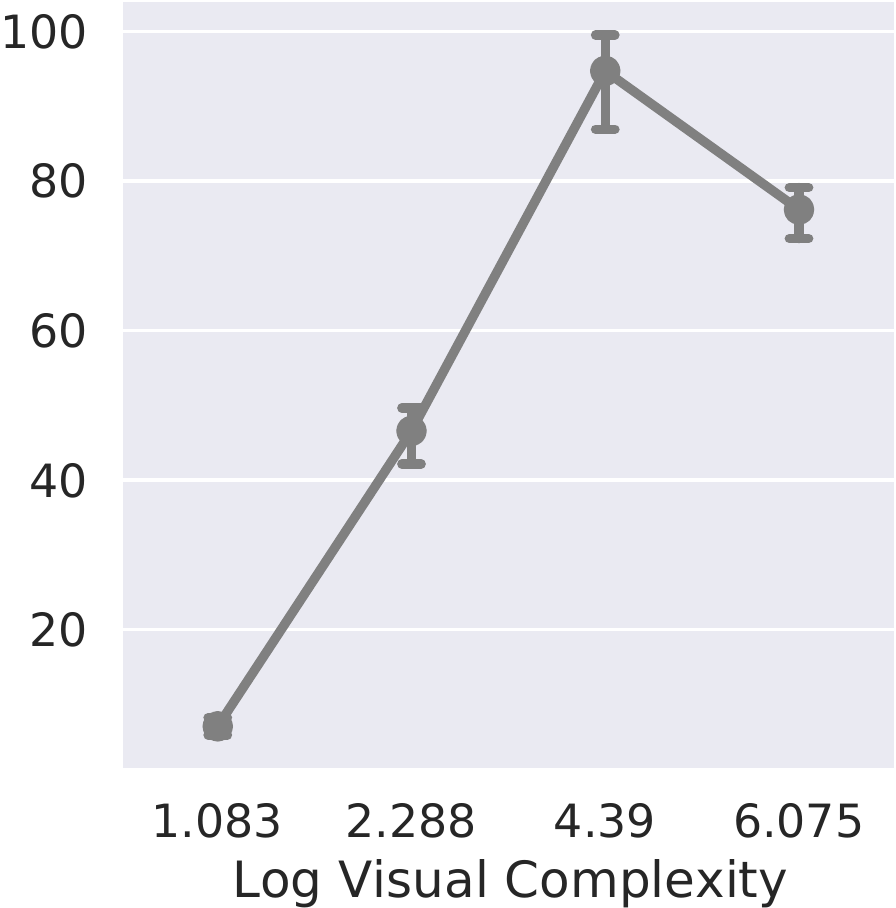}
        \caption{(d)}
        \label{fig:genaccp_r}
    \end{subfigure}
    \caption{\textbf{Quantitative results of \textit{Computation \vs Complexity}.} (a)(b) The rank correlation of similarity- and rule-based generalization with the four representations trained from four datasets. (c)(d) The rank correlation of similarity- and rule-based generalization according to the visual complexity. (L: LEGO, G:2D-Geo, A: ACRE, I: ImageNet) These plots reflect the landscape in \cref{fig:overview}.}
    \label{fig:baygenres}
\end{figure}

\begin{figure*}[bp]
    \centering
    \includegraphics[width=0.838\linewidth]{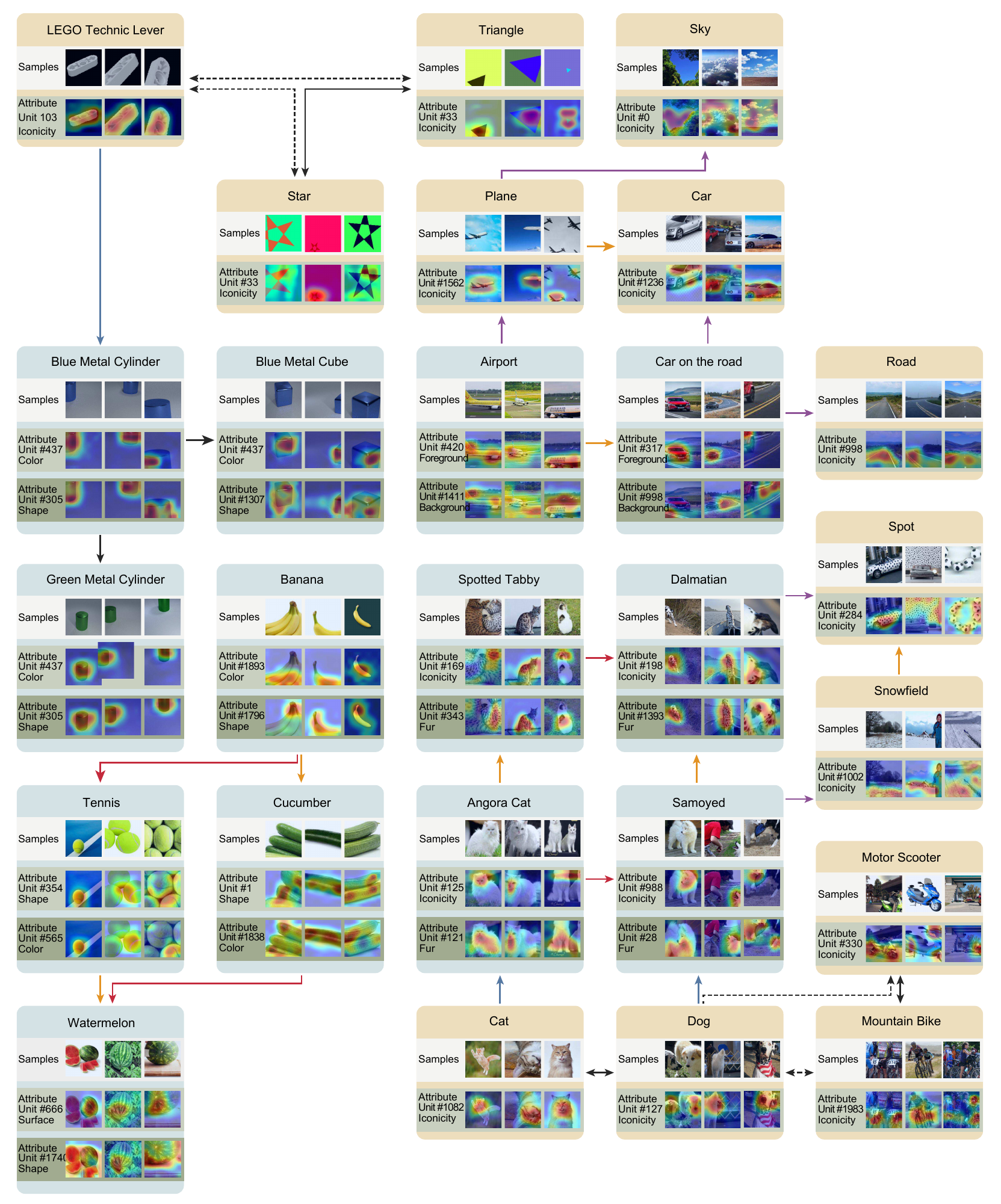}
    \caption{\textbf{A landscape of similarity- and rule-based generalization over concepts with relatively \textcolor{highcplx}{\textbf{high}} and \textcolor{lowcplx}{\textbf{low}} subjective complexity, considering both concept complexities and concept hierarchy.} \textbf{Bidirectional arrows} denote the similarity judgment between concepts, wherein concepts linked by solid lines are more similar than those linked by dashed lines. Arrows denote rules over concepts. Rule-based generalization in basic-level generalizes \textcolor{rulerd}{\textbf{given rules}} to \textcolor{ruleyl}{\textbf{unknown rules}}. Similarity shifts to rules when the sample hierarchy goes \textcolor{sup2sub}{\textbf{from superordinate-level to subordinate-level}} (\eg, from \textit{block} to \textit{blue cylinder}, from \textit{cat} to \textit{angora cat}). Rules shift to similarity as the sample hierarchy goes \textcolor{sub2sup}{\textbf{from subordinate-level to superordinate-level}} (\eg, from \textit{car on the road} to \textit{car}, from \textit{dalmatian} to \textit{spot}). We further notice a \textbf{confusing similarity judgment} between blue cylinder, blue cube, and green cylinder with distinct and shared attributes.}
    \label{fig:casestudy}
\end{figure*}

This experiment evaluates the capability of rule- and similarity-based generalization by the representations in \cref{sec:representation}. We predicted that under the same evaluation protocol, representations with relatively high subjective complexity outperform those with low subjective complexity in rule-based generalization, while the trend is the opposite in similarity-based generalization.

\paragraph{Method}

The evaluation of generalization is designed with two phases: in-domain and out-of-domain generalizations. The former consists of unseen samples from the test set of ACRE and ImageNet, whereas the latter contains unseen samples of unknown concepts collected from the internet. Each phase has a dataset with pairs for similarity-based generalization evaluation and a dataset with quadruples for rule-based generalization evaluation.

The evaluation protocol for similarity-based generalization extends its definition in \cref{sec:beyasian_generalization}. Formally, given unknown concept $c'$ and known concepts $c\in\mathcal{C}$, the ranking of the pairwise metric measurement is $S=\{\sigma_0(c_i,c')\geq \sigma_0(c_j,c')|c_i,c_j\in\mathcal{C}\}$. The representation ranking $S_r$ is obtained by the cosine similarity between two representation vectors $cos(z_i,z')$. The ground-truth ranking $S_h$ is obtained by human judgment. Hence, the generalization capability of the representation can be quantified through the rank correlation coefficient~\cite{spearman1961proof} as an accuracy measurement.

Similarly, the evaluation protocol for rule-based generalization is defined as follows. Given incomplete rule $r'(c_3,c')$ and known rules $r_i(c_1,c_2)\in\mathcal{R}$, the ranking score $R_r$ of representation is reduced to a cosine similarity calculation $cos(z_2-z_1+z_3,c')$, and the ground-truth ranking $R_h$ is obtained by human judgment~\cite{mikolov2013distributed}. We obtain the ground-truth concepts by literal meanings through the language representation model GloVe~\cite{pennington2014glove}. The image examples are retrieved from datasets (in-domain) or the internet (out-of-domain) with label embedding matching~\cite{vendrov2015order}.

\paragraph{Results}

The quantitative results for in-domain generalization evaluation are illustrated in \cref{fig:baygenres}. In similarity-based generalization, the representation trained from ImageNet outperforms others (over $15\%$), and LEGO outperforms its more complex counterparts 2D-Geo and ACRE (over $10\%$). In rule-based generalization, the representation trained from ACRE outperforms its more complex counterpart, ImageNet (over $20\%$). Though the models trained on ImageNet and ACRE reach the highest accuracy on similarity- and rule-based generalization, this is not likely due to over-fitting in training: The objective of visual categorization is different from that of generalization, thus the over-fitting on one visual categorization would not result in an over-fitting on other objectives. Intuitively, representations trained on more complex dataset span more complex attribute spaces. However, the result implies that the shift between similarity- and rule-based generalization is non-monotonic as the dataset complexity increases; it is more correlated to the subjective complexity based on \cref{sec:representation}. Hence, there is a significant negative relationship between the similarity-based generalization and the subjective complexity ($r=-.48$, $p<.05$), and a significant positive relationship between the rule-based generalization and the subjective complexity ($r=.68$, $p<.01$).

\begin{figure}[!h]
    \centering
    \includegraphics[width=0.8\linewidth]{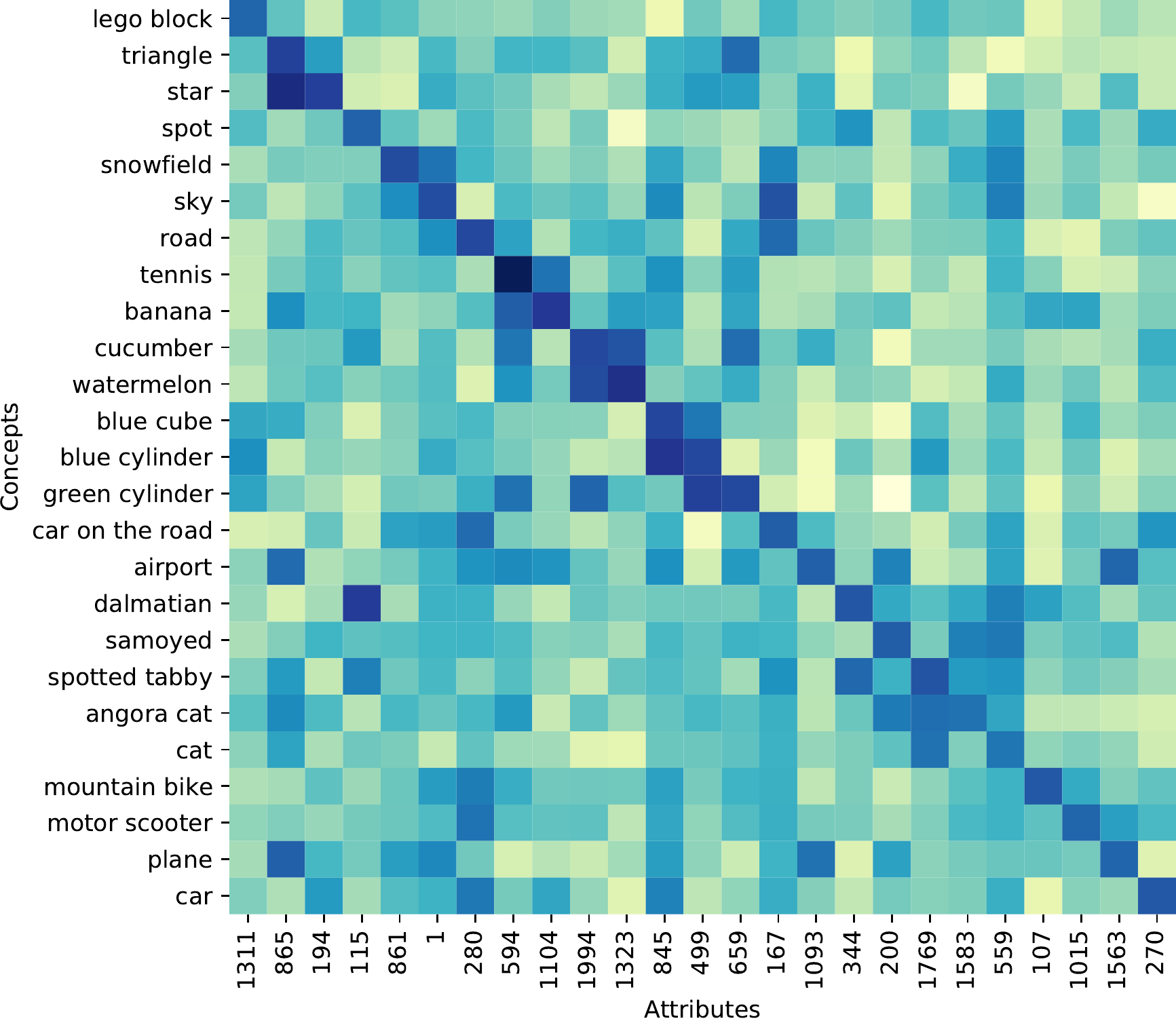}
    \caption{\textbf{The \ac{roa} matrix.} Most (21 out of 25) of the concepts are unknown; high saturation indicates high \ac{roa} value. The diagonal elements are the most representative attributes of all concepts.}
    \label{fig:ar_diag}
\end{figure}

\cref{fig:casestudy} illustrates the qualitative results for out-of-domain generalization. As shown in \cref{fig:ar_diag}, though never tuned on the unseen examples, the representation model also captures representative attributes for unknown concepts, which supports our argument in \cref{sec:beyasian_generalization} that \acs{roa} has the potential to serve as a prior for Bayesian generalization. Further, we visualize the most representative attributes of each concept by upsampling the activated feature vector to the size of the original image~\cite{bau2020understanding}; the attributes are located around the peaks. Most attributes with high \acs{roa} are explainable, such as the shape attribute shared by \emph{blue cylinder} and \emph{green cylinder}, shape and color captured by two distinct attributes in \emph{banana} and \emph{watermelon}, and foreground object (plane, car) and background (road, field) attributes in \emph{airport} and \emph{car on the road}. Those concepts with more than one meaningful attributes are sensitive to rule-based generalization. By contrast, those concepts with only one meaningful attribute, such as \emph{dog-like face} for \emph{dog}, \emph{car-like shape} for \emph{car}, are sensitive to similarity-based generalization.

\paragraph{Discussion}

The above experiment reveals that (i) both similarity- and rule-based generalizations are not significantly related to the visual complexity of datasets, (ii) the capability of similarity-based generalization has a significant negative relationship with the subjective complexity of representation, and (iii) the capability of rule-based generalization has a positive relationship with the subjective complexity of representation. We empirically articulate that the computation-mode-shift significantly exists, and similarity shifts to rules as the subjective complexity increases; please refer to the supplementary material for more details.

\subsection{A statistical interpretation}\label{sec:statinterp}

\paragraph{Subjective complexity in natural image statistics}

According to algorithmic information theory~\cite{chaitin1977algorithmic}, a concept's subjective complexity is proportional to the probability of perceiving this concept. This is consistent with the subjective complexity of visual concepts defined in our work. An attribute $z$ is representative for concept $c$ when $\ac{roa}(z,c)$ is relatively high; we have a high probability of observing the attribute by the concept (\eg, $P(z|c)=1$) or only by the concept (\ie, $\sum_{\hat{c}\neq c}P(\hat{c}|z)$ is small). Specifically, complex concepts (\eg, \textit{dog}, \textit{cat}), though consisting of many attributes (\eg, \textit{fur}, \textit{ear}), tend to have a unique attribute of \textit{view as a whole} to distinguish these concepts from others because we can hardly observe them in other concepts. Conversely, simple concepts (\eg, \textit{circle}, \textit{cylinder}) can be observed by many other concepts (\eg, \textit{wheel}, \textit{chimney}) and also have other attributes (\eg, \textit{number of angles}, \textit{smoothness}). Nevertheless, the attribute \textit{shape} is one of the simple attributes to describe these concepts; representation of these concepts emerges iconicity~\cite{guo2003towards,fay2010interactive,fay2013bootstrap,qiu2022emergent}.

Meanwhile, for those concepts that are either too simple or too complex (\eg, \textit{watermelon}, \textit{airport}), no unique or simple attribute can distinguish them from others; \ie, $\ac{roa}(z,c)$ is not high. In these cases, we have to describe them with more attributes. Of note, this interpretation is also in line with the principle of rational reference~\cite{frank2012predicting,goodman2016pragmatic}.

\paragraph{From similarity to rules}

Since similarity gradient can be viewed as a partial order defined on a single set~\cite{tenenbaum1999rules}, sorting hypotheses requires numerical comparison in the same domain. Hence, similarity judgment in a single attribute space $z_i$ is simply calculating the similarity between concepts $c_j$ and $c_k$ by $d(z_i^{(j)},z_i^{(k)})$, where $d(\cdot,\cdot)$ can be an arbitrary similarity or distance metric~\cite{ontanon2020overview}. As the number of independent attribute spaces increases (\ie, subjective complexity increases), the similarity becomes subtle as we have to consider multiple independent attributes. Of note, the attribute spaces are those obtained after dimension reduction~\cite{xie2016theory}. According to high-dimension geometry, those concept representations are almost distributed uniformly~\cite{blum2020foundations}, unless we assign weights to different attribute spaces by only considering very few attributes. For example, \textit{watermelon} is similar to \textit{tennis} in the attribute space of \textit{shape}, but it becomes \textit{cucumber} in the attribute space of \textit{color}; \textit{airport} is similar to \textit{plane} in the attribute space of \textit{foreground object} and is similar to \textit{land and sky} in the attribute space of \textit{background context}. In this work, we reduce similarity judgment over multiple attribute spaces to rules defining relations over two concepts: At least one shared attribute space bridges the two concepts.

\paragraph{From rules to similarities}

As the number of independent attribute spaces (\ie, subjective complexity) decreases, rules are moved back to similarity. For example, we have the rule relating \textit{dalmatian} to \textit{spotted tabby} by \textit{fur texture}, and can generalize it to \textit{samoyed to angora cat}. However, when the concepts are more complex (\eg, \textit{dalmatian} and \textit{samoyed} fall in \textit{dog}, or \textit{spotted tabby} and \textit{angora cat} belong to \textit{cat}), rules are difficult on these concepts; instead, we directly apply similarity judgment. 

\paragraph{Concept complexities and hierarchy}

When visual complexity moves from low to high, we have visual concepts move from \textit{simple and universal} to \textit{complex and unique}. We argue that these two ends consist of \textit{superordinate} concepts~\cite{xu2007word}, usually on higher hierarchies. Objects such as \textit{watermelon}, attribute-specified animals such as \textit{samoyed}, are subordinate concepts of \textit{ball} and \textit{dog}, respectively; scenes such as \textit{airport} are compositions of subordinate concepts like \textit{plane} and \textit{land and sky}. In a top-down view, we have concepts with increasing subjective complexity and more shared attribute spaces to generalize by rules. In a bottom-up view, the attribute spaces are reduced to the \textit{simple} or \textit{unique} ones, easy for similarity judgment.

\section{Conclusion}

We have analyzed the complexity of concept generalization in the natural visual world, in Marr's \textit{representational} and \textit{computational level}. At the representational level, the subjective complexities significantly fall in an inverted-U relation with the increment of visual complexity. At the computational level, the rule-based generalization is significantly positively correlated with the subjective complexity of the representation, while the trend is the opposite in similarity-based generalization. \acs{roa} bridges the two levels by unifying the frequentist properties of natural images (sensory-based) and the Bayesian properties of concepts (knowledge-derived)~\cite{bi2021dual}. It is easy to obtain, is flexible to an extent, and captures contextual rationality, thus may serve as humans' \emph{visual common sense}~\cite{zhu2020dark}. Please refer to the supplementary material for additional remarks.

The limitations of this work lead to several future directions:
(i) We only demonstrated the inverted-U relation and the correlation empirically. Can we provide them theoretically, from the aspect of information theory and statistics?
(ii) Can we further extend the generalization evaluation to a larger scale, that helps to probe the continuum space between similarity and rules quantitatively?
(iii) Are our findings consistent with those in other environments, where the concepts are represented in different modalities (\eg, language, audio, and tactile)?
(iv) If using only a few attributes with high \acs{roa} improves the accuracy of the visual categorization task, as \cref{sec:representation} suggests, can we build an algorithm that samples from \acs{roa} adaptively to task and data distribution for stronger generalization?
(v) If \acs{roa} reflects humans' visual commonsense, can we model the communications between individuals toward commonsense knowledge as a pursuit of the common grounds on representative attributes for the concepts to be communicated~\cite{tomasello2010origins}?
With many questions unanswered, we hope to shed light on future research on Bayesian generalization.

\paragraph{Acknowledgement}

Y.-Z. Shi, M. Xu, and Y. Zhu were supported in part by the Beijing Nova Program. Besides, we would like to thank Miss Chen Zhen (BIGAI) for making the nice figures in this paper. 

\paragraph{Reproducibility}

The source code for experiments in this work is available at \url{https://github.com/YuzheSHI/bayesian-generalization-complexity}.

{\small
\bibliographystyle{ieee_fullname}
\bibliography{references}
}

\clearpage 
\appendix

\section{Additional Remarks}\label{sec:remarks}

\subsection{The Uniqueness of the Natural Visual World}

Why do we only use the modality of vision to investigate the complexity of Bayesian generalization? Vision is unique for the diverse complexities both in the natural visual world and in the semantic space \cite{kersten2004object}, which relates vision to the discussion of levels of abstraction \cite{wu2008information}. To some extent, vision serves as the bridge between abstract language-derived knowledge and perceptual sensory-derived knowledge \cite{bi2021dual}. The two ends of the continuum of Bayesian generalization touch the functional essences of rule-based symbolic signals and similarity-based perceptual signals \cite{tenenbaum1999rules}. In particular, the very final development of symbolic signals leads to the emergence of language, based on the compositionality of symbols and rules as the basic feature of language. The psychology literature supports the hypothesis that language is emerged from visual communications by abstracting visual concepts toward hieroglyphs through their iconicity \cite{fay2010interactive,fay2013bootstrap,fay2014iconicity}. Both simple and universal visual concepts, such as geometric shapes, and complex and unique visual concepts, such as animals and artificial objects, are all related to corresponding abstract concepts by iconicity. By contrast, those concepts that are neither simple nor unique are unlikely to be abstracted by iconicity since they are described by multiple representative attributes---though each attribute can be generalized through iconicity respectively, putting different attribute spaces together is not making sense---by contrast, those concepts naturally satisfy the compositionality of language, thus are appropriate for rule-based generalization. In this sense, vision is not only a modality of data but is the hallmark of human intelligence, evolving perceptual sensory toward language for communications. Hence, vision is meaningful and sufficient for investigating the complexity of Bayesian generalization. 

Consider other modalities, say audio, the second common resource of sensory input. Although we could define audio complexity and try to correlate it with subjective complexity, audio is only a perceptual sensory---abstraction of raw audio is not related to any semantic meaning, thus does not provide much insight on human intelligence; also the diversity of audio complexity is far less than its visual counterpart. Hence, generalizing the experiments to audio data may be a bonus but never provides us insights as deep as that provided by visual data.

\subsection{The appropriateness of the computational modeling}

Thanks to Marr's paradigm~\cite{marr1982vision}, we could separate the computational-level problem and the representational-level problem, where we study computation problems regardless of their algorithmic representation or physical implementation in either humans or machines~\cite{lake2015human}. Hence, under the same computation problem, whether the algorithm is neural networks or brain circuits is not the problem in the scope.

Since the two parts of our computation problem---Bayesian generalization~\cite{tenenbaum2001generalization} and subjective complexity~\cite{lake2017building}---have established solid backgrounds in human cognition, we have a sufficient prerequisite for studying the complexities in the natural visual world. Though there may be infinite interpretations of human cognitive models~\cite{lieder2020resource}, constrained by previous theories and the principle of resource-rational analysis~\cite{lieder2020resource,gershman2015computational}, we can make assumptions about the Bayesian derivations. 

\section{Implementation Details}\label{sec:details}

\subsection{Implementing basic discriminative models}

The basic discriminative models are employed from ResNet \cite{he2016deep}, thus the feature space is spanned by a 512-d or 2048-d feature vector (dimensions are different by the different depths of ResNet architecture). All models are trained on eight NVIDIA A100 80GB GPUs.

\subsection{Implementing \acs{roa}}

In general, the \acs{roa} computes a score for each attribute $z_i$ over each concept $c$. The output of \acs{roa} is a matrix where the column space is the context of all the concepts in the natural visual world, and the row space is all the attributes. Assume we have three samples $\{x^{(1)},x^{(2)},x^{(3)}\}\in X$ of concept $c$, then the output of $f$ provides the attribute vectors $\mathbf{z}^{(1)},\mathbf{z}^{(2)},\mathbf{z}^{(3)}\in\mathbb{R}^{H\times W\times D}$ respectively. We then adaptively pool each feature map $\mathbf{z}^{(1)}_i,\mathbf{z}^{(2)}_i,\mathbf{z}^{(3)}_i\in\mathbb{R}^{H\times W\times d}$ in each dimension of the attribute vector to a scalar $z^{(1)}_i,z^{(2)}_i,z^{(3)}_i$, thus $\mathbf{z}^{(1)},\mathbf{z}^{(2)},\mathbf{z}^{(3)}\in\mathbb{R}^d$. $P(z_i|c)$ is calculated by normalizing over the dimensions of centroid vector of all $\mathbf{z}$ given the set of samples of concept $c$, \eg, $\widetilde{\bar{\mathbf{z}}^{(k)}},k=1,2,3$.

\subsection{Implementing Subjective Complexity Measurement}

Since the calculation of the absolute value of $L(\hat{c})$ may encounter multiple solutions, we employ accuracy gain \cite{bau2020understanding} to compute a relative $L(\hat{c})$ specifically for $\hat{c}$. The accuracy gain approach considers the categorization accuracy difference for a single concept before and after removing the effect of a specific neuron, defined as:
\begin{equation}
    \begin{aligned}
    \Delta Acc_K(\hat{c})=P\big(\hat{c}=c \big| c=\arg_c\max P(c|z_1,\dots,z_{K};\phi)\big) \\ -P\big(\hat{c}=c \big| c=\arg_c\max P(c|z_1,\dots,z_{K-1};\phi)\big),
    \end{aligned}
\end{equation}
where $K\geq 2$ and $\Delta Acc_1(\hat{c})=P\big(\hat{c}=c \big| c=\arg_c\max P(c|z_1;\phi)\big)$. Hence, the relative $L(\hat{c})$ is exactly computed by:
\begin{equation}
    L_{relative}(\hat{c})=\min_K\max \;Acc_K(c),
\end{equation}
which serves as the heuristic to search for the minimum $K$ to calculate absolute $L(\hat{c})$.

\section{Method Appropriateness Checking}
\subsection{Checking the assumptions of the two-lines test}\label{sec:two-lines}
The "two-lines test" requires a weaker assumption than the mostly used quadratic regression test for testing U-shapes \cite{lind2010or}, hence the former is employed instead of the latter. Let $y=f(x)$ be the ground-truth function, the U-shape assumes only a sign flip effect in discrete data, where there exists $x_c$ such that $f'(x), x\leq x_c$ and $f'(x), x\geq x_c$ has opposite signs \cite{lind2010or}. To note, since the data is originally discrete, there is no need to check the existence of $f'(x)$ because it is estimated based on the discrete data points. Hence, the basic hypothesis of the U-shape is that at least one such $x_c$ exists, and the null hypothesis is that no such $x_c$ exists. The null hypothesis is rejected by estimating many $x_c$ values and run two separate linear regressions for $x\leq x_c$ and $x\geq x_c$ respectively. The fact that two regression lines are of opposite sign rejects the null hypothesis. By contrast, the quadratic regression test assumes that the first-order derivative function $f'(x)$ is continuous in the domain. Hence, there is no need to employ the quadratic regression test.

\subsection{Checking the assumptions of the linear regression test}\label{sec:linear}

The assumptions of the test are (1) linearity of the data; (2) $x$ values are statistically independent; (3) the errors are homoscedastic and normally distributed. We did test the applicability of the linear regression test: (1) the two relations between rank correlation and subjective complexity are intuitively in lines ($[(0.1,7.8),(1.28,79.1),(2.91,46.7),(3.08,99.5)]$ and $[(0.1,17.1),(1.28,33.2),(2.91,15.8),(3.08,10.2)]$); (2) all the evaluations are run separately with different random seeds, thus the predictors are statistically independent; (3) since the only independent variable is the dataset, which is not likely to be the source of constant variance of the errors, the errors are homoscedastic. Consider the null hypothesis of the linear regression test that the coefficient $\beta_1$ is zero, which leads to a trivial solution. However, the p-values of both the positive and negative relations are less than 0.05, rejecting the null hypothesis.

\subsection{The correctness for combining representation and computation}\label{sec:articulate}

As illustrated in \cref{fig:baygenres}, we integrated the results in representation \vs complexity into this plot to use these plots to demonstrate the computation-mode-shift---the two U-shapes come with opposite trends intuitively show the landscape for concept complexity \vs the computation mode, that similarity-based generalization tends to emerge in concepts with very low or very high visual complexity (\ie, the concepts with low subjective complexity, on the left and right ends of the visual complexity axis), and rule-based generalization tends to emerge in concepts with neither very low nor very high visual complexity (\ie, the concepts with high subjective complexity, in the middle of the visual complexity axis). This is the exact claim of the paper. The quantitative results on the significant positive relation between rule-based generalization rank correlation and subjective complexity, and the significant negative relation between similarity-based generalization rank correlation and subjective complexity, both support the claim.
\section{Dataset Construction}
\subsection{Empirical Analysis Datasets}
\begin{figure*}[ht!]
    \centering
    \includegraphics[width=\linewidth]{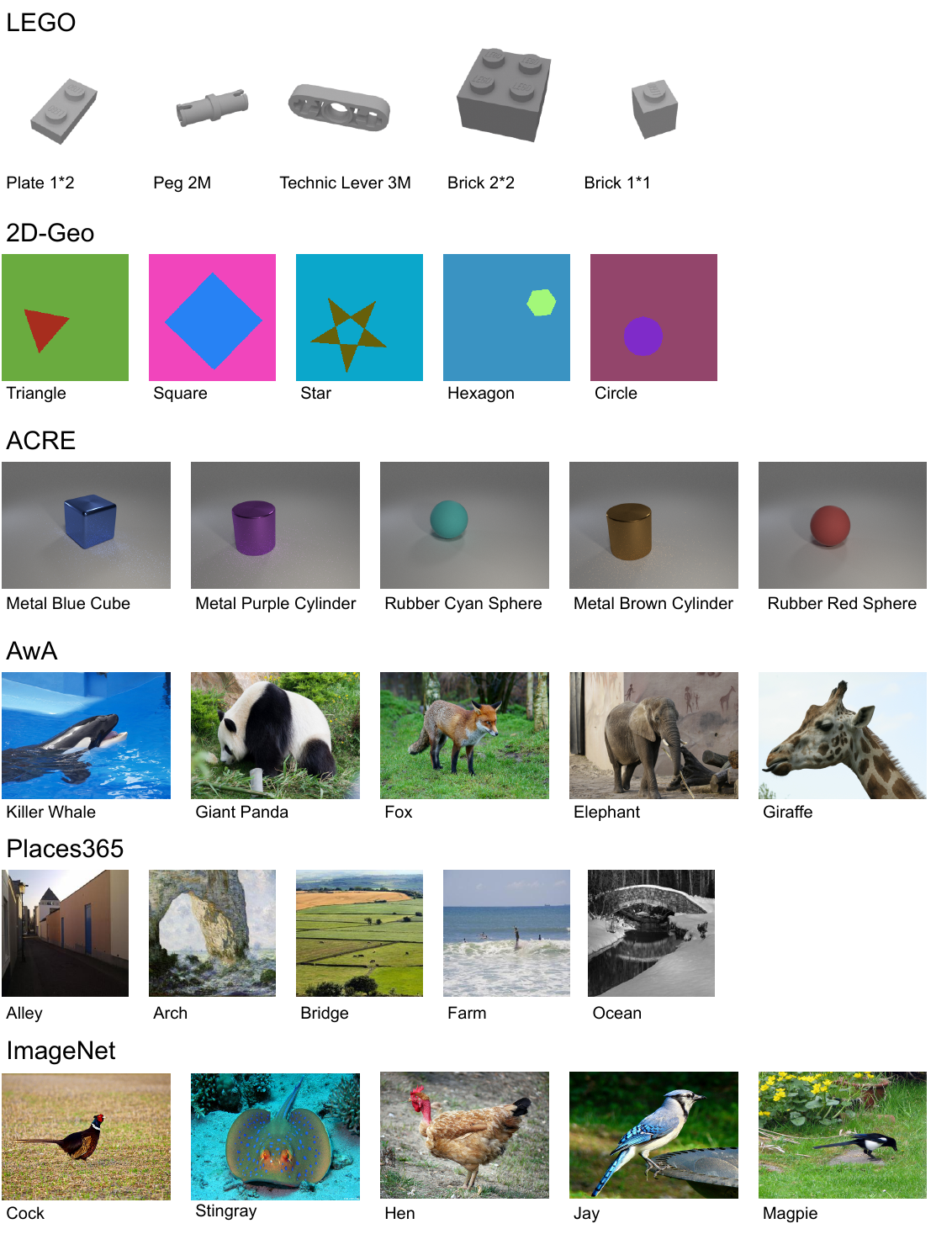}
    \caption{\textbf{Examples of datasets used in our work.}}
    \label{fig:dataset_example}
\end{figure*}

Several widely-used image datasets that represent different concept-wise visual complexity are selected: LEGO~\cite{tatman2017lego}, 2D-Geo~\cite{el20202d}, ACRE~\cite{zhang2021acre}, AwA~\cite{xian2018zero}, Places~\cite{zhou2017places}, and ImageNet~\cite{deng2009imagenet}. Especially, we use the Imagenet subset Imagenet-1k and the AwA2 version of AwA. The so-called ACRE dataset, although not officially released, is based on the well-known CLEVR universe~\cite{johnson2017clevr} and can be rendered with single object in one panel and without the blicket machine according to ~\cite{zhang2021acre}. See \Cref{fig:dataset_example} for some examples of the datasets we use. We limit images of each concept in all of these datasets to about 1k to ensure a balanced number of learning samples, which may lead to the gap between our models and the SOTA.

All the codes including the dataset construction, training and analyzing will be released. They are attached to this Supplementary for review.
\subsection{Definition of the Vocabulary}

We leverage a fully-connected probabilistic graph model to obtain the representativeness of every attribute for every concept, where each node is a piece of natural language that serves as either a concept or an attribute describing other concepts. We exploit the \acs{roa} in language to generate the in-domain and out-of-domain visual datasets for Bayesian generalization.
\begin{figure*}[ht!]
    \centering
    \includegraphics[width=\linewidth]{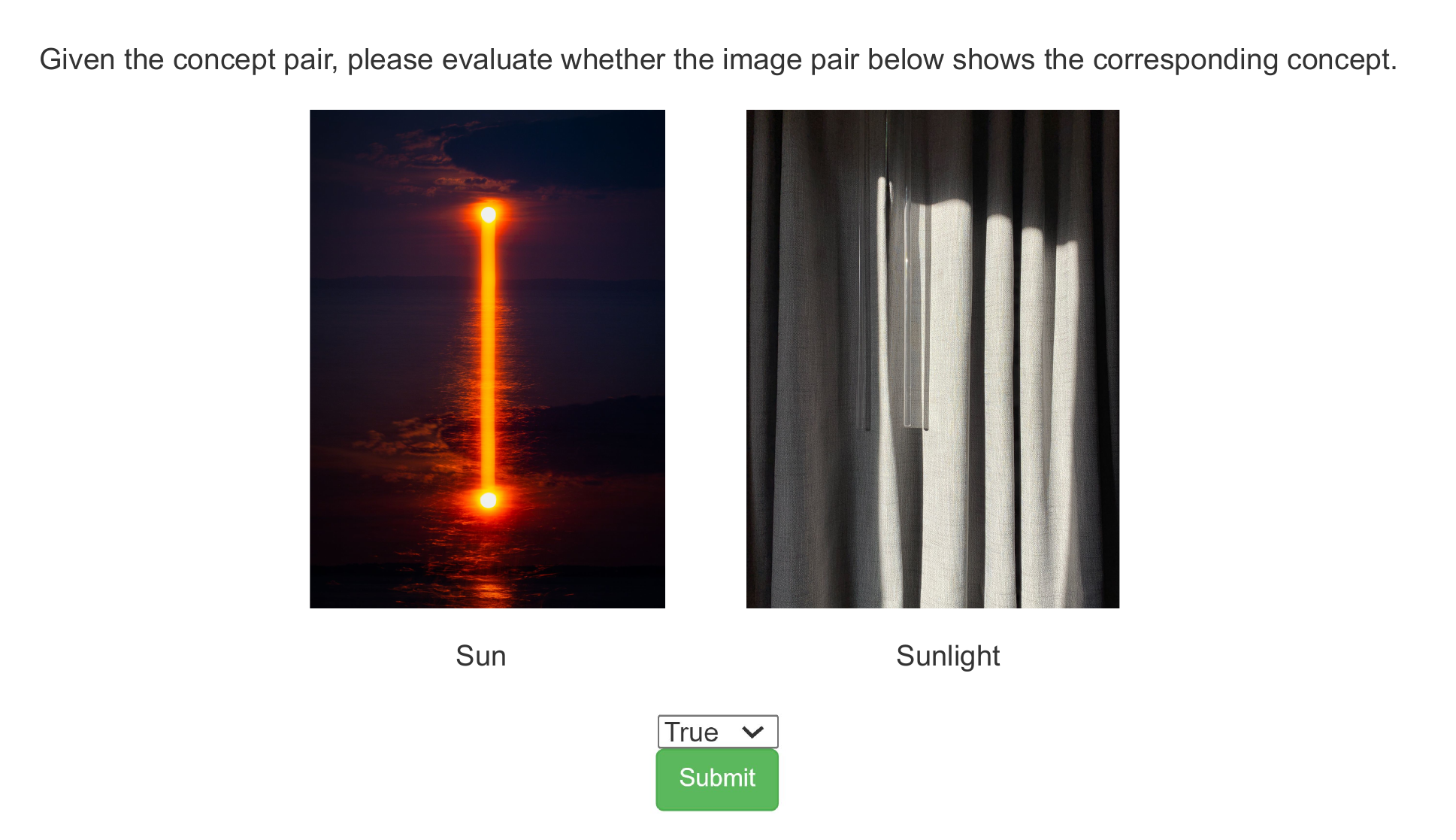}
    \caption{\textbf{The \ac{amt} interface used to collect human judgments.}}
    \label{fig:instruction_amt}
\end{figure*}
Technically, we use the vocabulary from a WordPiece model (e.g., the base version of Bert~\cite{vaswani2017attention}), where a word is tokenized into word pieces (also known as subwords) so that each word piece is an element of the dictionary. Non-word-initial units are prefixed with the sign ''\#\#'' as a continuation symbol. In this way, there is no Out-Of-Vocabulary. This brings the benefit of generalization over all words. Using all these words as attributes or features leads to sufficient coverage. Moreover, some symbols are reserved for unused placeholders, leaving room for features that the language cannot describe. The readers can refer to \texttt{vocab.txt} in the supplementary materials for more details about the attribute list.

\subsection{Human-in-the-loop dataset validation}

We constructed the \textit{similarity-based generalization} and the \textit{rule-based generalization} datasets using both manual approaches and automatic approaches. Details of all datasets are demonstrated in \cref{tab:dataset}.

For \textit{in-domain similarity-based generalization}, a concept pair with a human-annotated similarity score was first retrieved from MEN dataset \cite{bruni2014multimodal} and ImageNet dataset \cite{deng2009imagenet}. Next, we used \ac{amt} to crowd-source the image aligned to the concept. In total, 305 pairs were selected from 500 candidates. One image was aligned to each concept.

For \textit{in-domain rule-based generalization}, we generated dataset using objects of easy-to-disentangle attributes (\eg, shape and color) \cite{zhang2021acre,johnson2017clevr}. Based on these attributes, we constructed the quadruple relation (\eg, \texttt{blue cube:red cube::blue cylinder:red cylinder}). In total, 4800 images and 24 quadruple relations were collected.

For \textit{out-of-domain similarity-based generalization} and \textit{out-of-domain rule-based generalization}, we collected images from an open internet image dataset \cite{OpenImages2} based on a predefined set of similarity pairs and rule quadruples. Of note, all the selected pairs or rules were uniformly sampled from the dataset instead of manually picked. All the selected images were under human validation.

In the study, \ac{amt} workers recruited have acceptance rates higher than or equal to 90\% and approved hits more than 500. Each \ac{amt} worker was compensated at the rate of $0.01$ dollar per selection. In total, we have tested $1000$ judgments for $500$ concept pairs; two judgments per pair. \cref{fig:instruction_amt} shows an example of the \ac{amt} interface.

\begin{table*}[ht!]
    \centering
    \small
    \caption{\textbf{Details of the datasets for generalization evaluation.} Subordinate level indicates the concept being generalized to is a subordinate concept of the known ones, whereas superordinate level indicates the concept being generalized to is a superordinate concept of the known ones. Subordinate level, basic level, and superordinate level are terms introduced in \cite{xu2007word}.}
    \label{tab:dataset}
    \resizebox{\linewidth}{!}{%
        \begin{tabular}{c|cc|cccc}
            \hline
            Group               & \multicolumn{2}{c|}{In-domain}                      & \multicolumn{4}{c}{Out-of-domain}                                                                                                      \\ \hline
            Generalization type & \multicolumn{1}{c|}{Similarity-based} & Rule-based  & \multicolumn{1}{c|}{Similarity-based} & \multicolumn{3}{c}{Rule-based}                                                                 \\ \hline
            Concept hierarchy   & \multicolumn{1}{c|}{basic level}      & basic level & \multicolumn{1}{c|}{basic level}      & \multicolumn{1}{c|}{basic level} & \multicolumn{1}{c|}{subordinate level} & superordinate level \\ \hline
            Test-set size       & \multicolumn{1}{c|}{305}              & 24          & \multicolumn{1}{c|}{21}               & \multicolumn{1}{c|}{10}          & \multicolumn{1}{c|}{10}                & 10                  \\ \hline
        \end{tabular}%
    }%
\end{table*}

\section{Additional Results}

\subsection{The Convergence of Representation \texorpdfstring{\vs}{} Generalization}

\begin{figure*}[ht!]
    \centering
    \includegraphics[width=0.8\linewidth]{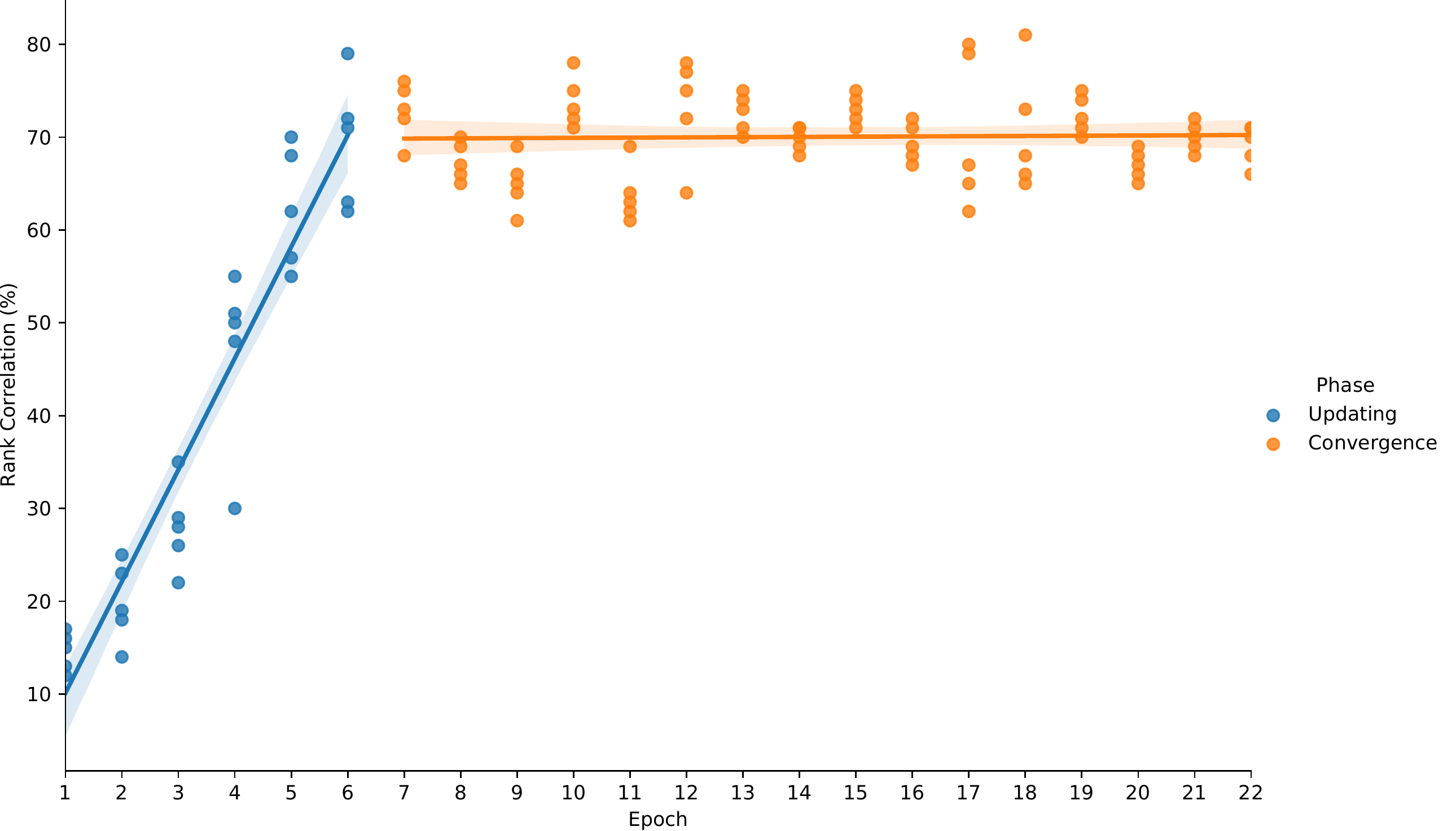}
    \caption{\textbf{Rank correlation of generalization \wrt the number of training epochs for visual categorization.}}
    \label{fig:convergence}
\end{figure*}

Does the training setting of the representation model affect its generalization ability? \cref{fig:convergence} shows the rank correlation on in-domain generalization evaluation \wrt the number of training epochs for visual categorization. This result empirically shows that the generalization ability converges when the representation models are well trained after 6-10 epochs, and that ability is stable after convergence. The \textcolor{orange}{regression line} is significantly vertical to the y-axis ($b=.03,a=69.67,p<1e-4$). Hence, we can assume that there are no significant distinctions of generalization ability between representation models being trained to convergence but with different training settings.

\subsection{Additional Visualization Results of \acs{roa}}

\begin{sidewaysfigure*}[ht!]
    \centering
    \begin{subfigure}[ht!]{\linewidth}
        \centering
        \includegraphics[width=\linewidth]{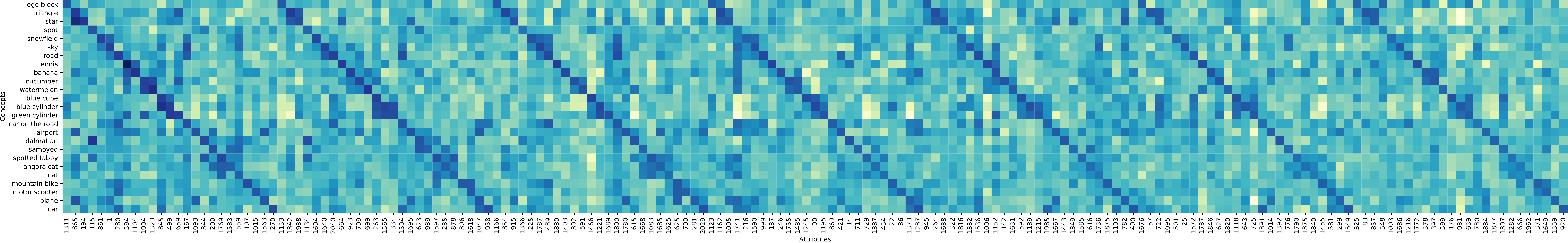}
        \caption{\textbf{\acs{roa} top-7 confusion matrices.}}
        \label{fig:roa_bigdiag}
    \end{subfigure}   
    \begin{subfigure}[ht!]{\linewidth}
        \centering
        \includegraphics[width=\linewidth]{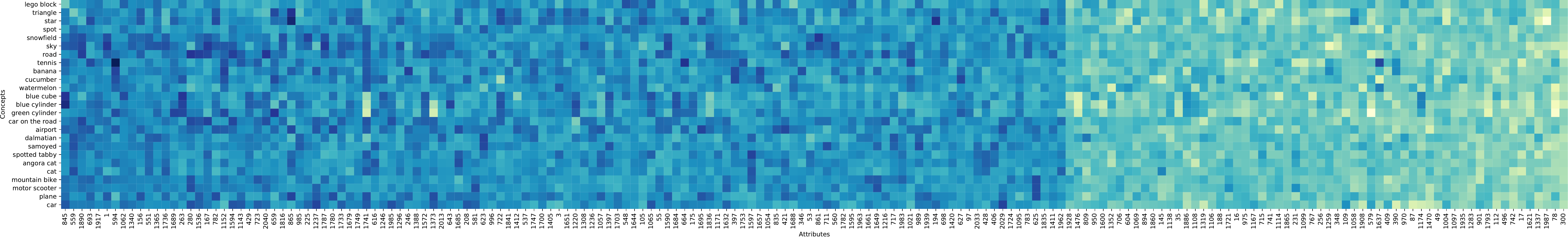}
        \caption{\textbf{\acs{roa} with decreasing mean.}}
        \label{fig:roa_mean_dec}
    \end{subfigure}
    \begin{subfigure}[ht!]{\linewidth}
        \centering
        \includegraphics[width=\linewidth]{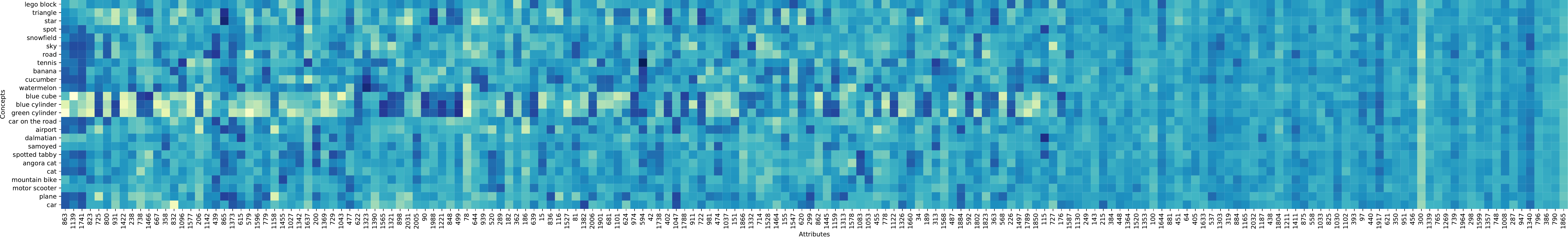}
        \caption{\textbf{\acs{roa} with decreasing variance.}}
        \label{fig:roa_var_dec}
    \end{subfigure}
    \caption{\textbf{Additional visualizations of \acsp{roa}.}}
    \label{fig:extroa}
\end{sidewaysfigure*}

Additional visualization results of \acs{roa} are illustrated in \cref{fig:extroa}. Most (21 out of 25) concepts are unknown; high saturation indicates high \ac{roa} value. 

\cref{fig:roa_bigdiag} shows the concatenation of the 7 confusion matrices where the $n$-th diagonal indicates the $n$-top \acs{roa} of the concepts.

\cref{fig:roa_mean_dec} shows the concatenation of 120 highest (from the left) and 60 lowest (from the right) attributes with the mean of \acs{roa} in the context.

\cref{fig:roa_var_dec} shows the concatenation of 120 highest (from the left) and 60 lowest (from the right) attributes with the variance of \acs{roa} in the context.

\clearpage

\end{document}